\documentclass[11pt]{article}

\usepackage[final]{acl}

\makeatletter
\@ifpackageloaded{lineno}{%
  \AddToHook{shipout/after}{\@LN@col{1}}%
}{}
\makeatother

\usepackage{times}
\usepackage{latexsym}

\usepackage[T1]{fontenc}

\usepackage[utf8]{inputenc}

\usepackage{microtype}

\usepackage{amsmath}

\usepackage{algorithm}
\usepackage{algpseudocode}

\usepackage{inconsolata}

\usepackage{graphicx}

\usepackage{booktabs}

\usepackage{float}
\usepackage{placeins}
\usepackage{dblfloatfix}

\usepackage{url}

\usepackage{listings}
\usepackage{xcolor}
\usepackage{tcolorbox}
\tcbuselibrary{listings,breakable}

\newtcolorbox{promptbox}[1][]{
  colback=cyan!8,
  colframe=cyan!60!black,
  fonttitle=\bfseries\small,
  title=#1,
  breakable,
  before skip=0.6em,
  after skip=0.6em,
  boxrule=0.5pt,
  left=4pt,
  right=4pt,
  top=2pt,
  bottom=2pt
}

\newtcolorbox{casebox}[1][]{
  colback=orange!8,
  colframe=orange!60!black,
  fonttitle=\bfseries\small,
  title=#1,
  breakable,
  before skip=0.6em,
  after skip=0.6em,
  boxrule=0.5pt,
  left=4pt,
  right=4pt,
  top=2pt,
  bottom=2pt
}

\title{BMAM: Brain-inspired Multi-Agent Memory Framework}

\author{
  Yang Li\textsuperscript{1} \quad
  Jiaxiang Liu\textsuperscript{1} \quad
  Yusong Wang\textsuperscript{2} \quad
  Yujie Wu\textsuperscript{3} \quad
  Mingkun Xu\textsuperscript{1}\thanks{Corresponding author.} \\
  \textsuperscript{1}Guangdong Institute of Intelligence Science and Technology, Zhuhai, China \\
  \textsuperscript{2}Institute of Science Tokyo \quad
  \textsuperscript{3}The Hong Kong Polytechnic University \\[2pt]
  \texttt{\{liyang,\,liujiaxiang,\,xumingkun\}@gdiist.cn} \\
  \texttt{wangyi@lr.pi.titech.ac.jp} \quad
  \texttt{yu-jie.wu@polyu.edu.hk}
}

\begin{document}
\maketitle


\begin{abstract}
Language-model-based agents operating over extended interaction horizons face persistent challenges in preserving temporally grounded information and maintaining behavioral consistency across sessions, a failure mode we term \emph{soul erosion}. We present \textbf{BMAM} (Brain-inspired Multi-Agent Memory), a general-purpose memory architecture that models agent memory as a set of functionally specialized subsystems rather than a single unstructured store. Inspired by cognitive memory systems, BMAM decomposes memory into episodic, semantic, salience-aware, and control-oriented components that operate at complementary time scales, organised as a six-phase memory lifecycle. To support long-horizon reasoning, BMAM organises episodic memories along explicit timelines and retrieves evidence by fusing multiple complementary signals. Experiments on the \textbf{LoCoMo} benchmark show that BMAM achieves \textbf{78.45\% accuracy}, outperforming seven memory-augmented baselines. Pairwise ablations reveal \emph{super-additive} synergy between brain-region components rather than redundant stacking, and a Soul Portability Test demonstrates \textbf{87.5\% identity-integrity} across full memory export, clear, and restore. A targeted refinement of the temporal-trigger heuristics raises LongMemEval multi-session accuracy from $45.2\%$ to $56.4\%$, validating the architectural decomposition behind BMAM.Code is available at \url{https://github.com/innovation64/BMAM}. 
\end{abstract}
\section{Introduction}

Language-model-based agents increasingly operate in settings that require maintaining and reasoning over information accumulated across extended interactions, spanning diverse tasks, domains, and time scales. Such agents must retain past experiences, organize them into usable memory structures, and retrieve relevant information under varying goals and contexts. However, large language models are constrained by finite context windows and lack an explicit mechanism for managing long-term memory beyond the current input~\citep{MemGPT,LoCoMo}. Retrieval-augmented generation (RAG) partially alleviates this limitation by fetching external documents on demand, but it treats memory as an external text repository rather than an internal, evolving system. As a result, RAG-style approaches provide limited support for persistent memory accumulation, temporal organization, and cross-episode reasoning, motivating the need for a general-purpose memory framework that can support long-horizon agent behavior across tasks rather than task-specific retrieval pipelines~\citep{MemorySurvey2024,MemorySurvey2025}.

Evidence from cognitive science suggests that memory is not a single monolithic store, but is supported by multiple functionally specialized subsystems operating over complementary time scales (e.g., fast episodic encoding alongside slower semantic consolidation and executive control) \citep{ComplementaryLearningSystemsTheory}. Inspired by this view, we propose \textbf{BMAM} (Brain-inspired Multi-Agent Memory Framework), a brain-inspired multi-agent memory architecture that decomposes agent memory into interacting subsystems responsible for episodic storage, semantic consolidation, salience-aware selection, and intent-conditioned control \citep{li2025condambigqa}. BMAM constructs internal memory representations rather than relying solely on external retrieval, and employs a timeline-indexed episodic memory organization to support temporally grounded access to past experiences. The framework further integrates a hybrid retrieval mechanism that combines lexical, dense, knowledge-graph, and temporal signals via reciprocal rank fusion, together with asynchronous memory consolidation processes inspired by complementary learning principles. To coordinate memory access across different temporal scales, BMAM adopts a hierarchical memory control mechanism from recent work, enabling both fast context-level access and slower consolidated memory retrieval. In preliminary analyses of long-horizon agent behavior, we observe a recurring failure pattern in which fragmented or misaligned memory leads to degradation in temporal coherence and identity-related behavior across interactions, which we refer to as {soul erosion}, providing a diagnostic lens for failures of long-term memory management in general-purpose agent settings.
Our main contributions are:



\begin{itemize}
  \item We identify and characterize \textbf{soul erosion}, a recurring failure pattern in long-horizon agent behavior where fragmented or misaligned memory leads to degradation in temporal coherence and identity-related behavior.
  \item We propose \textbf{BMAM}, a brain-inspired framework that addresses this challenge by decomposing memory into specialized subsystems (episodic, semantic, salience). Crucially, we introduce a \textbf{timeline-indexed organization} and a \textbf{hybrid retrieval strategy} that fuses lexical, semantic, and temporal signals for robust grounding.
  \item We validate BMAM on the \textbf{LoCoMo} benchmark, achieving \textbf{78.45\%} accuracy and outperforming baselines in long-horizon settings. Further ablation studies empirically confirm the critical role of the hippocampus-inspired subsystem in enabling temporal reasoning.
\end{itemize}

\begin{figure*}[!t]
\centering
\includegraphics[width=0.88\textwidth]{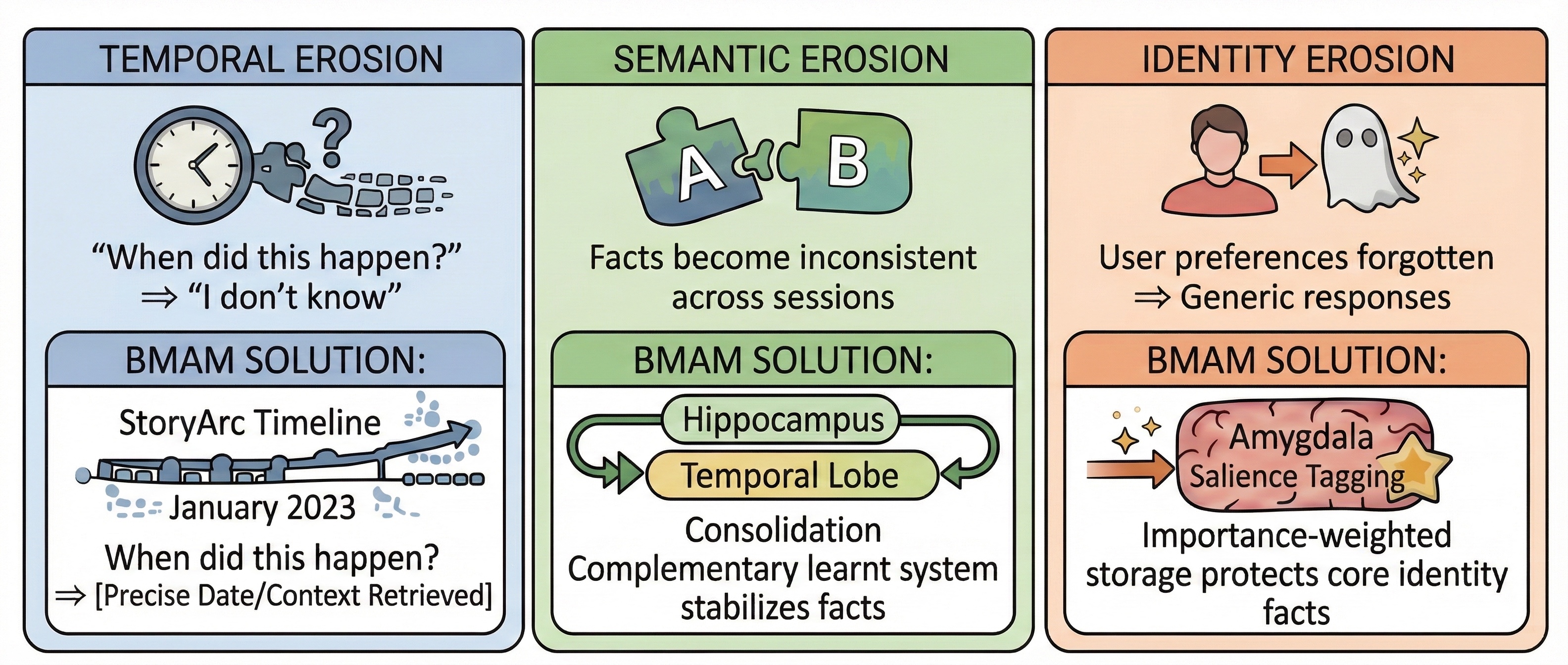}
\caption{Soul erosion types and BMAM countermeasures. Each erosion mechanism requires a specialized defense: temporal erosion is addressed by StoryArc timeline indexing, semantic erosion by hippocampus-to-temporal-lobe consolidation, and identity erosion by amygdala salience tagging.}
\label{fig:soul-erosion}
\end{figure*}

\paragraph{Soul Erosion: Why Memory Matters}
\label{sec:soul-erosion}

We use the term \emph{soul erosion} to describe a recurring failure pattern in long-horizon agent interactions, where fragmented or misaligned memory leads to degradation in behavioral continuity and identity-related behavior. Analogous to how human identity relies on the continuity of autobiographical memory~\citep{wilson2003identity,bluck2013therefore}, an AI agent's ``soul'' (its consistent preferences, behavioral tendencies, and interaction patterns) may gradually degrade when long-term memory is poorly organized or inconsistently accessed.

\paragraph{Formal Definition}
We formalize soul erosion as a composite degradation metric over three orthogonal dimensions. Let $\mathcal{M}_t$ denote the agent's memory state at interaction step $t$. We define the \textbf{soulfulness score} $\mathcal{S}$ as:
\begin{equation}
\mathcal{S}(\mathcal{M}_t) = \alpha \cdot T(\mathcal{M}_t) + \beta \cdot C(\mathcal{M}_t) + \gamma \cdot I(\mathcal{M}_t)
\label{eq:soulfulness}
\end{equation}
where $T(\cdot)$ measures \emph{temporal coherence} (ability to correctly order and recall when events occurred), $C(\cdot)$ measures \emph{semantic consistency} (absence of factual contradictions), and $I(\cdot)$ measures \emph{identity preservation} (retention of user-specific preferences and traits). The weights $\alpha, \beta, \gamma \geq 0$ with $\alpha + \beta + \gamma = 1$ reflect task-specific importance.

\textbf{Soul erosion} is then defined as the degradation of soulfulness over time:
\begin{equation}
\mathcal{E}(t_0, t) = \mathcal{S}(\mathcal{M}_{t_0}) - \mathcal{S}(\mathcal{M}_t)
\label{eq:erosion}
\end{equation}
where $t_0$ is a reference point (e.g., initial interaction or last memory consolidation). A positive $\mathcal{E}$ indicates soul erosion has occurred. In our experiments, we operationalize these components using benchmark proxies: $T$ via LoCoMo temporal accuracy, $C$ via cross-session consistency metrics, and $I$ via PrefEval and PersonaMem scores.

\textbf{Soul erosion encompasses three distinct failure modes} (Figure~\ref{fig:soul-erosion}), each arising from different memory failures and requiring specialized countermeasures:

\paragraph{(1) Temporal Erosion}
The agent loses track of \emph{when} events occurred, leading to anachronistic or temporally inconsistent responses. Cognitive research shows that temporal context is fundamental to episodic memory organization~\citep{HowardKahana2002,EichenbaumTimeAndMemory}, and benchmarks like LoCoMo and LongMemEval~\citep{LoCoMo,LongMemEval} reveal that LLM agents frequently fail on temporal queries. As shown in Figure~\ref{fig:soul-erosion} (left), without explicit temporal organization, the agent may confuse event order, overlook durations, or fail to answer time-dependent queries. BMAM addresses temporal erosion through StoryArc timeline indexing, which maintains explicit temporal structure over stored experiences.

\paragraph{(2) Semantic Erosion}
Facts and relationships degrade or become internally inconsistent across interactions. This mirrors the forgetting and interference phenomena studied in human memory~\citep{WixstedForgetting,AndersonInhibition}, where memories compete and degrade without proper consolidation. As depicted in Figure~\ref{fig:soul-erosion} (center), the agent may provide contradictory answers about the same entity over time. HippoRAG~\citep{HippoRAG} and memory surveys~\citep{MemorySurvey2024} highlight this challenge. BMAM counters semantic erosion through hippocampus-to-temporal-lobe consolidation, which promotes frequently accessed and high-confidence episodic memories into stable semantic representations.

\paragraph{(3) Identity Erosion}
User preferences, personality traits, and persistent behavioral patterns may be overwritten or lost as new context accumulates. Research on autobiographical memory emphasizes that identity coherence depends on preserving self-relevant experiences~\citep{ConwayIdentity,McAdamsNarrative}. Benchmarks like PersonaMem and PrefEval~\citep{PersonaMem,PrefEval} demonstrate that current systems struggle to maintain user-specific information. As shown in Figure~\ref{fig:soul-erosion} (right), this failure mode undermines personalization: the agent ``forgets'' who the user is. BMAM mitigates identity erosion through amygdala-inspired salience tagging, which prioritizes identity-relevant information and protects it from being overwhelmed by transient context.

\paragraph{Multi-Agent Coordination as Erosion Defense}
A central design insight of BMAM is that these three forms of erosion arise from distinct memory failures and cannot be fully addressed by a single mechanism. Cognitive neuroscience research demonstrates that human memory relies on multiple specialized systems (the hippocampus for episodic encoding, the neocortex for semantic consolidation, and the amygdala for emotional salience) that interact to maintain coherent long-term memory~\citep{ComplementaryLearningSystemsTheory}. Inspired by this functional specialization, BMAM distributes memory functions across multiple interacting components, each targeting a specific erosion type (Figure~\ref{fig:soul-erosion}). Our ablation studies (Table~\ref{tab:brain-region-ablation}) empirically validate this design: removing the hippocampus-inspired episodic memory causes the largest performance drop, confirming its critical role, while other components contribute complementary defenses against different erosion types.

\section{Background and Related Work}

\paragraph{Memory Architectures for LLM Agents}
RAG improves factual grounding but treats memory as implicit and transient; retrieved passages are not reorganized into stable internal structures~\citep{DHRAG,MRAG,ActiveRAG}. Agent-centric frameworks address this limitation through explicit memory management. MemGPT pioneered virtual context management by treating LLMs as operating systems with hierarchical memory tiers~\citep{MemGPT}. MemoryBank extends this with forgetting mechanisms inspired by Ebbinghaus curves~\citep{MemoryBank}. A-MEM introduces agentic memory that autonomously manages storage and retrieval~\citep{AMem}. Production systems like Mem0, Memobase, and MemOS provide scalable memory APIs with multi-component stores~\citep{Mem0,Memobase,MemOS}. Hierarchical approaches organize memory by semantic abstraction levels~\citep{HMEM,HRM,MemGAS}. Memory-augmented transformers have explored various mechanisms for extending context, including segment-level recurrence \citep{dai-etal-2019-transformer}, kNN-augmented attention \citep{wu2022memorizing}, and brain-inspired episodic memory 
\citep{Larimar}. These approaches primarily target token-level or sequence-level prediction, whereas BMAM targets long-horizon agent memory management: what to store, how to organize it temporally, and how to retrieve it under changing goals.

\paragraph{Brain-Inspired and Cognitive Approaches}
Cognitive neuroscience motivates separating fast episodic encoding from slower semantic consolidation and salience-based prioritization~\citep{ComplementaryLearningSystemsTheory}. This principle has inspired systems like HippoRAG for hippocampus-style indexing~\citep{HippoRAG}, Nemori for event segmentation~\citep{Nemori}, and reflective memory systems that learn from experience through prospective and retrospective reflection~\citep{RMM,Reflexion}. Recent architectures emphasize tight coupling between perception and memory, forming closed loops that support adaptive long-term memory~\citep{Voyager,GenerativeAgents}. Compared to these approaches, BMAM differs in three key aspects: (1) \emph{multi-region coordination}: while HippoRAG focuses on hippocampal pattern separation, BMAM models interactions among multiple brain-region analogs (hippocampus, temporal lobe, amygdala, prefrontal cortex); (2) \emph{explicit temporal indexing}: unlike Nemori's event boundaries, BMAM maintains continuous timeline structures that support arbitrary temporal queries; (3) \emph{salience-aware consolidation}: BMAM integrates amygdala-inspired importance signals into the consolidation process, prioritizing identity-relevant information over transient context.

\paragraph{Benchmarks}
Long-term memory benchmarks evaluate temporal reasoning (LoCoMo~\citep{LoCoMo}, LongMemEval~\citep{LongMemEval}), preference consistency (PrefEval~\citep{PrefEval}), and persona recall (PersonaMem~\citep{PersonaMem}), providing complementary perspectives on the challenges BMAM addresses.

\begin{figure*}[!t]
\centering
\includegraphics[width=0.88\textwidth]{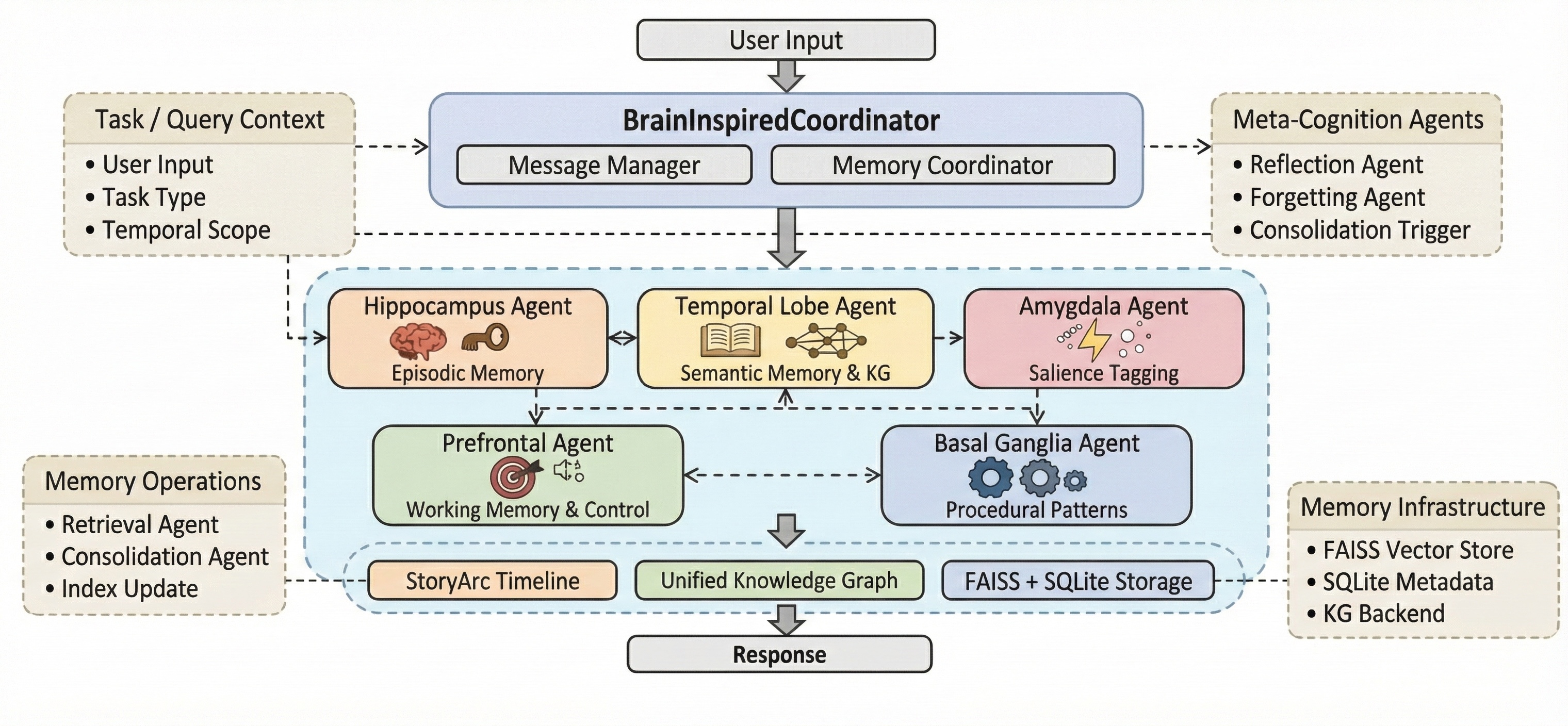}
\caption{BMAM architecture overview. A central coordinator orchestrates multiple functionally specialized memory subsystems sharing a unified memory substrate with episodic timelines, a knowledge graph, and vector-based storage.}
\label{fig:bmam-architecture}
\end{figure*}

\section{BMAM Framework}

\label{sec:bmam}

BMAM adopts a coordinator-centered multi-agent architecture that decomposes long-term memory into functionally specialized components while maintaining a unified memory substrate. A central coordinator routes information among interacting subsystems responsible for memory storage, retrieval, consolidation, and control, enabling modular specialization without fragmenting memory state.
\begin{algorithm}[!t]
\scriptsize  
\caption{BMAM Memory Lifecycle Pipeline}
\label{alg:bmam-pipeline}
\begin{algorithmic}[1]
\Require query $q$, context $C$, regions $\mathcal{B}=\{\text{Hip},\text{Tmp},\text{Amg},\text{Pfc},\text{Bg}\}$
\Ensure response $r$
\Statex \textit{\# Phase 1: Query analysis \& routing}
\State $f \gets \textsc{AnalyzeFeatures}(q, C)$ \Comment{temporal/entity/intent}
\State $\langle b_1,\ldots,b_K\rangle, s \gets \textsc{LearnableRouter}(q)$
\State $A \gets \textsc{ThalamusGating}(q, C)$ \Comment{$A[b]\!\in\!\{0,1\}$}
\Statex \textit{\# Phase 2: Region-specific parallel access}
\ForAll{activated region $b \in \mathcal{B}$ \textbf{in parallel}}
  \State $M_{\text{hip}} \gets \text{Hip}.\textsc{EpisodicSearch}(q)$
  \State $M_{\text{tmp}} \gets \text{Tmp}.\textsc{SemanticSearch}(q)$ \Comment{BM25\,+\,vec\,+\,KG}
  \State $M_{\text{amg}} \gets \text{Amg}.\textsc{SalienceRetrieve}(q)$
  \State $M_{\text{pfc}} \gets \text{Pfc}.\textsc{WMLookup}(q)$
  \State $M_{\text{bg}}  \gets \text{Bg}.\textsc{ProceduralMatch}(q)$
\EndFor
\Statex \textit{\# Phase 3: Cross-region resonance scoring}
\State $M \gets M_{\text{hip}}\cup M_{\text{tmp}}\cup M_{\text{amg}}\cup M_{\text{pfc}}\cup M_{\text{bg}}$
\ForAll{$m \in M$}
  \State $m.\rho \gets |\{b\!:\,m\in M_b\}|$ \Comment{resonance count}
  \State $m.\sigma \gets \sum_i w_i \cdot \phi_i(m,q)$ \Comment{weighted score}
\EndFor
\Statex \textit{\# Phase 4: Hippocampal--Prefrontal loop}
\State $i \gets 0$
\Repeat
  \State $M \gets \text{Amg}.\textsc{ModulateSalience}(M, q)$
  \State $M \gets M \cup \text{Tmp}.\textsc{FillSemanticGaps}(q, M)$
  \State $\gamma, \mathit{gaps} \gets \text{Pfc}.\textsc{EvaluateEvidence}(q, M)$
  \If{$\gamma < \tau$}
    \State $q' \gets \text{Pfc}.\textsc{GapAnalysis}(q, \mathit{gaps})$
    \State $M \gets M \cup \text{Hip}.\textsc{ExpandedRetrieval}(q')$
  \EndIf
  \State $i \gets i+1$
\Until{$\gamma \geq \tau$ \textbf{or} $i = I_{\max}$}
\Statex \textit{\# Phase 5: Fusion, reranking, structured reasoning}
\State $M^{\star} \gets \textsc{LLMRerank}(\textsc{TopN}(M,\sigma), q)$
\If{$\textsc{KGTrigger}(q)$} $M^{\star} \gets M^{\star}\cup \text{Tmp}.\textsc{KGQuery}(q)$ \EndIf
\If{$\textsc{TempQuery}(q)$} $r_t \gets \textsc{TemporalReason}(q, M^{\star})$ \EndIf
\Statex \textit{\# Phase 6: Generation \& memory write-back}
\State $r \gets \textsc{Generate}(q, M^{\star}, C)$
\State $\text{Hip}.\textsc{EncodeEpisode}(q, r, C)$
\State $\text{Amg}.\textsc{TagEmotion}(q, r)$
\If{$\text{importance}>0.6 \land \text{access}>2$}
  \State $\text{Tmp}.\textsc{Consolidate}(\text{Hip}.\text{episodes})$
\EndIf
\State $\text{Pfc}.\textsc{UpdateWM}(q, r)$
\State $\textsc{FeedbackLoop}.\textsc{UpdateVectors}(M^{\star}, \gamma)$
\State \Return $r$
\end{algorithmic}
\end{algorithm}
\paragraph{Memory Loop and Coordination}
BMAM implements an explicit memory loop inspired by hippocampus--neocortex dynamics. Incoming experiences are encoded into episodic memory using fast, discriminative representations and tagged with salience signals, while relevant content is maintained in a constrained working-memory buffer to support immediate reasoning. Over time, selected episodic information is consolidated into semantic memory and a shared knowledge graph. Retrieval closes the loop by jointly accessing episodic and semantic evidence under temporal constraints, with feedback signals adjusting consolidation priorities and routing decisions.

\paragraph{Functionally Specialized Memory Components}
BMAM decomposes memory into complementary subsystems with explicit roles and capacities. Episodic memory stores temporally grounded interaction traces and supports discriminative addressing. Semantic memory consolidates stable facts and relations into a shared knowledge graph. A salience-aware component computes importance signals from interaction cues (e.g., novelty, conflict, or user feedback) that modulate consolidation scheduling and retrieval weighting.

The \textbf{Prefrontal} component implements executive control functions inspired by the prefrontal cortex's role in working memory maintenance and cognitive control~\citep{MillerCohen2001}. Specifically, it performs three functions: (1) \emph{query routing}, classifying incoming queries along dimensions (temporal, identity, preference, factual) to determine which memory subsystems to consult; (2) \emph{working-memory buffering}, maintaining a capacity-limited buffer (10 items) of recent context for immediate reasoning without full memory retrieval; and (3) \emph{attention allocation}, dynamically weighting evidence sources based on query requirements. Control-oriented components, including the Prefrontal buffer and Basal Ganglia procedural patterns, together provide complementary protections against different forms of memory degradation.

\paragraph{Unified Memory Substrate and Temporal Indexing}
BMAM employs a unified memory substrate that combines key--value episodic storage, vector-based similarity indexing, and a shared knowledge graph. Episodic memories are organized into a timeline-indexed structure that records minimal narrative units indexed by entities, events, and timestamps. This temporal organization enables queries involving order, duration, and temporal relations (e.g., before/after, first/last), while consolidation processes selectively lift episodic information into semantic form to ensure consistency across representations.

\paragraph{Hierarchical Coordination and Retrieval}
To support long-horizon interactions, BMAM adopts hierarchical memory coordination mechanisms that regulate memory access and updates across multiple time scales. Fast paths support immediate context-level access, while slower paths govern semantic consolidation and procedural stabilization. Retrieval integrates fast-path detection, iterative interaction between episodic and control components, and uncertainty-driven multi-round retrieval. Evidence from episodic memory, semantic memory, and the knowledge graph is combined with temporal constraints, and feedback signals dynamically reweight lexical, dense, entity-based, and temporal cues.

\paragraph{Memory Lifecycle}
BMAM models memory as a dynamic lifecycle governing encoding, consolidation, retrieval, and revision, organised as a six-phase pipeline (Algorithm~\ref{alg:bmam-pipeline}) and summarised diagrammatically in Appendix Figure~\ref{fig:memory-lifecycle}.

\paragraph{Input Analysis and Episodic Encoding}
We first analyze each incoming interaction to extract entities, temporal expressions, and intent cues relevant to memory formation. The interaction is encoded as an episodic memory trace, capturing the contextual content together with inferred temporal and semantic attributes. Salience signals are computed from interaction cues (e.g., novelty, conflict, or user feedback) and attached to the episode. To support efficient short-term reasoning, a compact summary of recent episodes is maintained in a constrained working-memory buffer.

\paragraph{Consolidation and Temporal Organization}
Next, BMAM employs a complementary learning process in which frequently accessed and high-confidence episodic memories are selectively consolidated into semantic memory. Consolidated information populates a shared knowledge graph that maintains stable facts and relations across interactions. In parallel, episodic memories are organized into a timeline-indexed structure that records entity-centric events with associated temporal information. This temporal organization enables reasoning over event order, relative timing, and durations, supporting queries such as \emph{when}, \emph{before/after}, and \emph{how long} without requiring full episodic recall.
\paragraph{Hybrid Retrieval and Temporally Grounded Answering}
To answer a query, BMAM retrieves relevant evidence from multiple sources, including episodic memory, semantic memory, and the timeline-indexed event structure. Each source $s \in \mathcal{S}$ produces a ranked list of candidates, and lexical, dense, relational, and temporal signals are fused using (weighted) reciprocal rank fusion:
\begin{equation}
\mathrm{score}(d \mid q) = \sum_{s \in \mathcal{S}} \frac{w_s}{k + \mathrm{rank}_s(d \mid q)},
\label{eq:wrrf}
\end{equation}
where $\mathrm{rank}_s(d \mid q)$ is the rank of candidate $d$ under source $s$, $k=60$ is the smoothing constant following standard RRF practice, and $w_s$ reflects the current preference over evidence sources. For time-dependent questions, temporal evidence is extracted from the timeline organization to compute relative orderings and durations, which are then used to generate temporally grounded answers. This retrieval process is adaptive: uncertainty and salience signals may trigger additional retrieval rounds or reweight evidence sources.

\paragraph{Background Optimization and Memory Revision}
In parallel with online interaction, memory organization in BMAM is continuously refined through background processes. Episodic memories may be reconsolidated when re-accessed, increasing their stability or updating their content as new evidence emerges. Low-value or outdated memories are gradually pruned, while salience-relevant episodes receive prioritized consolidation. These processes allow BMAM to revise memory over time, preventing uncontrolled growth and reducing the accumulation of inconsistent or obsolete information.

\paragraph{Computational Complexity}
Local processing (routing, BM25/FAISS retrieval, resonance scoring) runs in tens of milliseconds; the hippocampal--prefrontal loop is bounded by $I_{\max}=3$. End-to-end latency is dominated by LLM calls: \textbf{3--5 per query} (2--3 on simple queries; 4--5 when Pfc triggers re-retrieval). Measured on LoCoMo across eight runs, BMAM reaches \textbf{17.3\,s/query} best and \textbf{22.4\,s/query} average, at $\sim$\$0.001/query using \texttt{gpt-4o-mini}---comparable to rerank-RAG (2--3 calls) on simple queries, with additional cost concentrated on multi-hop queries, where BMAM's margin over baselines is largest (Appendix Table~\ref{tab:per-category-ablation}).

\paragraph{Continual Learning and Plasticity}
Over longer interaction horizons, BMAM treats memory as a plastic substrate rather than a static store. Continual learning emerges from ongoing consolidation and reconsolidation, whereby retrieved evidence can update semantic memory instead of being frozen after first storage. Conceptually, if $p_t(f)$ denotes the confidence of a semantic fact $f$ at time $t$, and $\hat{p}_t(f)$ is an evidence-based estimate from new retrieval/verification, then memory revision can be expressed as an exponential moving average:
\begin{equation}
p_{t+1}(f) = (1-\lambda)p_t(f) + \lambda \hat{p}_t(f),
\label{eq:conf-update}
\end{equation}
where $\lambda \in (0,1)$ is the update rate. This enables knowledge updates while damping noisy evidence. When confidence is low or information is incomplete, the system may actively seek clarification through follow-up interaction, strengthening memory traces and reducing uncertainty. Over time, adaptive routing, salience-weighted storage, and confidence-calibrated retrieval (e.g., by adjusting $w_s$ in Eq.~\ref{eq:wrrf}) change what is stored, how it is indexed, and how evidence is combined, enabling BMAM to evolve its memory behavior as experience accumulates.

\FloatBarrier
\section{Experiments}
\label{sec:experiments}

\subsection{Experimental Setup}

We evaluate BMAM on four benchmarks designed to test long-horizon memory and personalization capabilities (Table~\ref{tab:bmam-datasets}). Our evaluation focuses primarily on \textbf{LoCoMo} and \textbf{LongMemEval}, which together capture complementary challenges in conversational memory, temporal reasoning, and memory consistency over extended interactions.

\paragraph{Primary Benchmarks}
We focus primarily on \textbf{LoCoMo}~\citep{LoCoMo} and \textbf{LongMemEval}~\citep{LongMemEval}, which together capture complementary challenges in long-horizon memory. LoCoMo evaluates recall of facts, relationships, and events across extended multi-session dialogues, emphasizing single-hop factual recall, multi-hop reasoning, and temporally grounded questions. LongMemEval complements this with cross-session recall, preference tracking, knowledge updates, and explicit temporal reasoning, stressing memory consistency under evolving information.
\paragraph{Additional Benchmarks}
We further evaluate BMAM on \textbf{PersonaMem} and \textbf{PrefEval}, which focus on persona consistency and preference alignment, respectively. These benchmarks test whether memory systems can preserve user-specific information and behavioral preferences across interactions, complementing the conversational and temporal challenges posed by LoCoMo and LongMemEval.

\begin{table}[htbp]
\centering
{\small
\setlength{\tabcolsep}{4pt}
\resizebox{\linewidth}{!}{%
\begin{tabular}{p{0.22\columnwidth}p{0.28\columnwidth}p{0.26\columnwidth}p{0.18\columnwidth}}
\toprule
Dataset & Scale & Task Focus & Metric \\
\midrule
LoCoMo & 10 groups, 1986 QA & long-horizon dialogue & Accuracy \\
LongMemEval & 500 questions & long-term memory & Accuracy \\
PersonaMem & 20 users, 589 QA & persona recall (MCQ) & Accuracy \\
PrefEval & 1000 questions & preference alignment & Pers. rate \\
\bottomrule
\end{tabular}
}}
\caption{Datasets and evaluation metrics used in BMAM experiments.}
\label{tab:bmam-datasets}
\end{table}
\begin{table*}[t]
\centering
\small
\setlength{\tabcolsep}{6pt}
\begin{tabular}{lccccc}
\toprule
Method & Single-hop & Multi-hop & Temporal & Open-domain & Overall \\
\midrule
\textbf{BMAM (ours)} & \textbf{82.00} & \textbf{70.42} & 62.31 & 79.55 & \textbf{78.45} \\
MemOS-1031$^{\dagger}$ & 64.54 & 57.29 & 71.34 & 79.90 & 73.90 \\
Memobase   & 73.12 & 64.65 & \textbf{81.20} & 53.12 & 72.01 \\
Mem0       & 73.33 & 58.75 & 52.54 & 45.83 & 64.57 \\
MIRIX      & 68.32 & 54.26 & 68.54 & 46.88 & 64.33 \\
Zep        & 65.23 & 52.12 & 54.82 & 33.33 & 59.22 \\
\bottomrule
\end{tabular}
\caption{Head-to-head LoCoMo accuracy per category against seven memory-augmented baselines. BMAM leads the re-run MemOS baseline by 4.55 points overall, 8.67 on single-hop, and 5.77 on multi-hop. Memobase is stronger on temporal-only queries; this is addressed by the trigger refinement (Table~\ref{tab:locomo-postfix}). $^{\dagger}$Re-run with \texttt{gpt-4o-mini} for strict comparability; other baselines use reported numbers.}
\label{tab:locomo-baselines}
\end{table*}

\paragraph{Baselines}
We compare BMAM against seven memory-augmented LLM systems:
\textbf{MemOS}~\citep{MemOS}, a memory operating system with unified memory scheduling;
\textbf{Mem0}~\citep{Mem0}, a scalable memory-centric architecture with optional graph-based memory;
\textbf{MIRIX}~\citep{MIRIX}, a multi-agent system with six specialized memory types;
\textbf{Zep}~\citep{Zep}, a temporally-aware knowledge graph engine;
\textbf{Memobase}~\citep{Memobase}, \textbf{Supermemory}~\citep{Supermemory}, and \textbf{MemU}~\citep{MemU}.
Baseline results are from~\citet{MemOS}; we re-run MemOS with GPT-4o-mini for fair comparison.

\paragraph{Evaluation Protocol}
For all benchmarks, persistent memory is reset between independent evaluation units (e.g., LoCoMo conversation groups or individual users) while being preserved within each unit to reflect realistic interaction histories. Conversation logs are ingested through BMAM's memory lifecycle prior to evaluation, and queries are issued in evaluation mode without additional learning. We follow the evaluation protocol, metrics, and judge prompts from MemOS\footnote{Official repository: \url{https://github.com/MemTensor/MemOS}; paper: \url{https://arxiv.org/abs/2507.03724}};
baselines are evaluated using their official scripts. Crucially, to ensure a fair comparison, we re-evaluated the strongest baseline (MemOS) using the identical LLM backend (GPT-4o-mini) as BMAM, eliminating discrepancies arising from model version updates.
During evaluation, all background processes (consolidation, reconsolidation, pruning) are disabled; memory state is frozen after ingestion to prevent test-time learning.

\subsection{Results}
\label{sec:results}

\begin{table}[htbp]
\centering
{\small
\setlength{\tabcolsep}{4pt}
\begin{tabular}{p{0.24\columnwidth}p{0.20\columnwidth}p{0.16\columnwidth}p{0.24\columnwidth}}
\toprule
Dataset & Metric & Score & Correct/Total \\
\midrule
LoCoMo & Accuracy & \textbf{78.45\%} & 1558/1986 \\
LongMemEval & Accuracy & 67.60\% & 338/500 \\
PersonaMem & Accuracy & 48.9\% & 288/589 \\
PrefEval & Pers. rate & 72.90\% & 729/1000 \\
\bottomrule
\end{tabular}
}
\caption{BMAM results across four long-term memory benchmarks.}
\label{tab:bmam-results}
\end{table}

Across benchmarks, BMAM is strongest on LoCoMo and PrefEval, while PersonaMem remains challenging: its multiple-choice format requires exact surface-form matching, whereas BMAM's retrieval is optimized for open-ended generation, and its emphasis on shallow persona attributes differs from BMAM's focus on temporally grounded identity. We provide a more detailed discussion in Appendix~\ref{sec:baseline-comparisons}. Temporal reasoning remains a key open challenge, and improving normalized temporal outputs and cross-session integration is an important direction for future work.Table~\ref{tab:bmam-results} summarizes our main results.

\paragraph{Temporal-Trigger Refinement}
An audit of the temporal reasoning module showed that broad trigger heuristics (bare keywords \textit{before}, \textit{after}, \textit{when}) and a low confidence threshold ($0.35$) caused over-firing on non-temporal queries. Three targeted fixes---narrowing triggers to precise forms (\textit{when did}, \textit{how long ago}), raising the threshold to $0.55$, and expanding the non-temporal filter set from 4 to 30$+$ patterns---raise temporal accuracy from $76.9\%$ to $\mathbf{84.6\%}$ and yield $\mathbf{78.89\%}$ overall on a single LoCoMo group ($N{=}199$), consistent with the 10-group main result ($78.45\%$); full per-category numbers are in Appendix Table~\ref{tab:locomo-postfix}.

\paragraph{LoCoMo Performance}
On LoCoMo, BMAM achieves $\mathbf{78.45\%}$ overall under our MemOS-aligned evaluation protocol. Table~\ref{tab:locomo-baselines} reports the full head-to-head against seven memory-augmented baselines across all five question types; because other baselines use reported numbers under possibly different backends, we re-ran MemOS with \texttt{gpt-4o-mini} for strict comparability. Performance varies by question type: single-hop ($82.0\%$), multi-hop ($70.4\%$), temporal ($62.3\%$), and open-domain ($79.6\%$). BMAM leads on every category except temporal, where Memobase is stronger; we address this weakness via the trigger refinement below.

\paragraph{LongMemEval Performance}
BMAM reaches $67.60\%$ overall on LongMemEval, excelling within single sessions ($100\%$ preference; $87.1\%$ user facts) and on knowledge updates ($70.5\%$), with the main gaps on cross-session integration ($52.6\%$) and temporal reasoning ($59.4\%$); full per-category results are in Appendix Table~\ref{tab:longmemeval-breakdown}.

\paragraph{Post-Refinement Re-run on LongMemEval ($N{=}500$)}
After the trigger fix, re-running the full LongMemEval benchmark raises multi-session accuracy from $45.2\%$ to $\mathbf{56.4\%}$, while temporal reasoning drops from $59.4\%$ to $54.9\%$ (a precision--recall trade-off discussed in Limitations). The four other categories are unaffected, and overall accuracy is stable ($67.6\% \to 66.8\%$).The per-category breakdown is reported in Appendix Table~\ref{tab:longmemeval-rerun}.

\paragraph{Soul Portability: Empirical Verification of Identity Persistence}
To directly test whether BMAM's decomposition supports portable identity, we ran a four-phase test: (1)~\emph{shape} 19 sessions (603 memories) and record baseline answers to 20 identity-probing questions; (2)~\emph{export} complete memory state to a \texttt{.bma} archive; (3)~wipe state, \emph{restore} from archive (1{,}206 entries across 4 regions); (4)~re-answer the 20 questions and compute similarity against the baseline. The composite \textbf{Soul Integrity Score of 87.5\% (Grade~A)}---$75\%$ semantic equivalence and $60\%$ exact match (Table~\ref{tab:soul-portability})---provides quantitative evidence that accumulated identity can be exported, transferred, and restored with high fidelity, validating the soul-erosion framework beyond benchmark accuracy.

\begin{table}[h]
\centering
\small
\begin{tabular}{lc}
\toprule
Metric & Value \\
\midrule
Memories shaped (19 sessions) & 603 \\
Brain regions exported & 5\,/\,5 \\
Memory entries restored & 1{,}206 \\
Brain regions restored & 4\,/\,5 \\
Exact match (20 questions) & 12\,/\,20 (60.0\,\%) \\
Semantic match ($\geq 0.8$) & 15\,/\,20 (75.0\,\%) \\
\midrule
\textbf{Soul Integrity Score} & \textbf{87.5\,\% (Grade A)} \\
\bottomrule
\end{tabular}
\caption{Soul Portability Test results. The system can export, clear, and restore identity with 75\,\% semantic equivalence across an independent question set.}
\label{tab:soul-portability}
\end{table}

\begin{figure}[t]
\centering
\includegraphics[width=0.95\columnwidth]{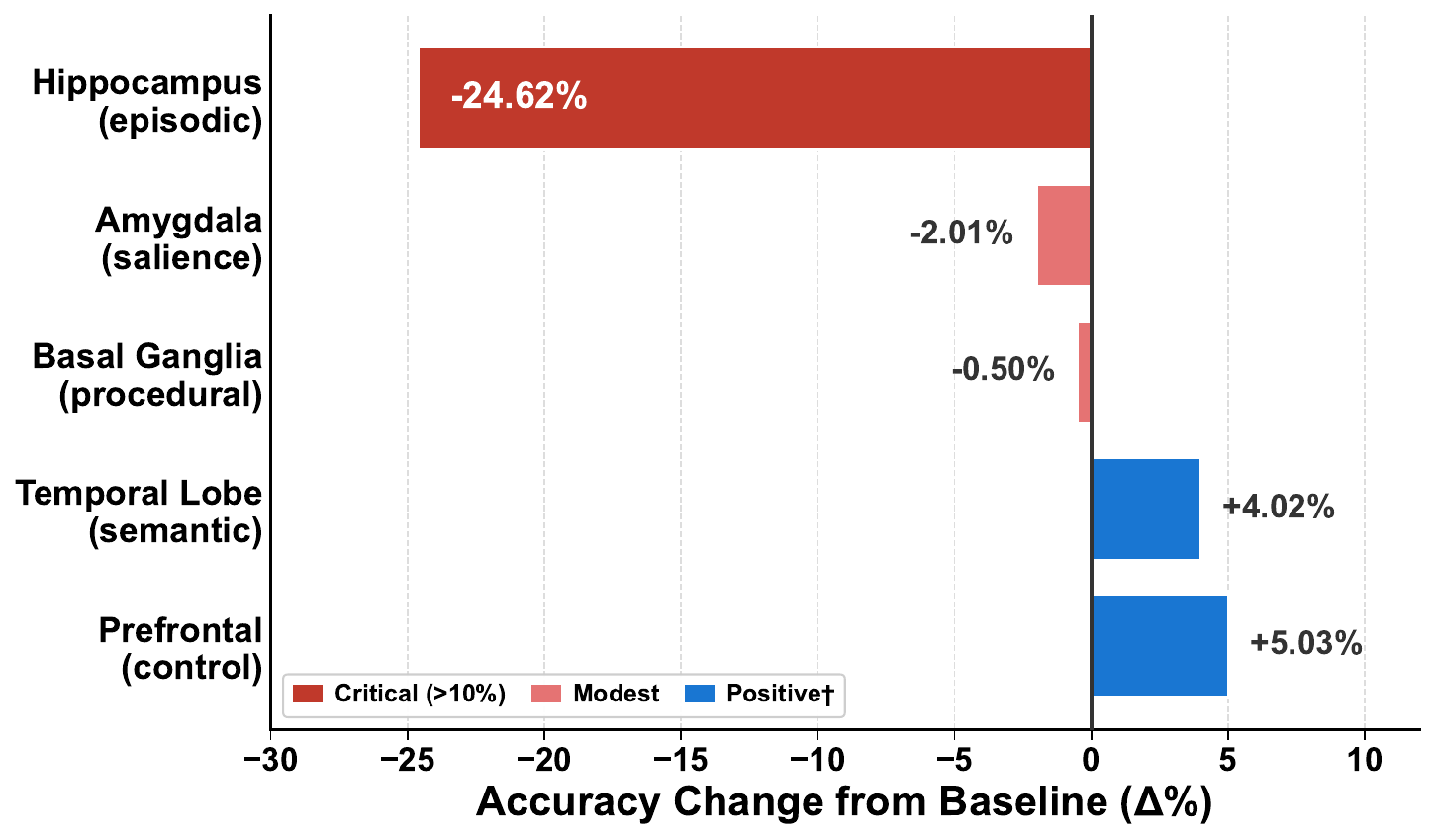}
\caption{Brain-region ablation on LoCoMo. Hippocampus removal causes a $24.62\%$ drop, validating episodic memory as the critical backbone. Positive deltas for Prefrontal and Temporal Lobe are unpacked in Table~\ref{tab:per-category-ablation} (per-category breakdown).}
\label{fig:ablation-analysis}
\end{figure}

\paragraph{Ablation Analysis}
To examine component contributions, we conduct ablation experiments on a LoCoMo subset (Figure~\ref{fig:ablation-analysis}). Removing the hippocampus-inspired episodic memory leads to a $24.62\%$ accuracy drop, confirming its central role.

\paragraph{Per-Category Ablation: Where Each Region Matters}
Disaggregating the ablation by query category (Table~\ref{tab:per-category-ablation}) reveals a clear specialisation pattern that the aggregate numbers mask: removing Prefrontal or Temporal Lobe raises open-domain accuracy by 11 and 13 points but drops multi-hop accuracy by 16 and 14 points respectively; removing Hippocampus collapses multi-hop accuracy from $73.0\%$ to $27.0\%$. Higher-order regions thus trade easy-query routing overhead for the working memory and semantic support that multi-hop reasoning genuinely requires.

\begin{table}[t]
\centering
\small
\setlength{\tabcolsep}{3pt}
\begin{tabular}{lcccc}
\toprule
Configuration & Overall & M-hop & Open-d. & Adv. \\
              &         & ($n{=}37$) & ($n{=}70$) & ($n{=}47$) \\
\midrule
Full BMAM         & 77.39 & \textbf{72.97} & 78.57 & 85.11 \\
w/o Hippocampus   & 52.76 & 27.03 & 50.00 & 89.36 \\
w/o Prefrontal    & 82.41 & 56.76 & 90.00 & 93.62 \\
w/o Temporal Lobe & 81.41 & 59.46 & 91.43 & 93.62 \\
w/o Amygdala      & 75.38 & 59.46 & 82.86 & 87.23 \\
w/o Basal Ganglia & 76.88 & 72.97 & 82.86 & 76.60 \\
\bottomrule
\end{tabular}
\caption{Per-category brain-region ablation on LoCoMo ($N{=}199$). Compare ablated rows to Full BMAM to read the drop; multi-hop is the category where each higher-order region matters most.}
\label{tab:per-category-ablation}
\end{table}

\paragraph{Component Synergy: Pairwise Ablation}
To test whether the brain regions are merely additive or interact synergistically, we ran pairwise ablations that disable two regions simultaneously (Table~\ref{tab:pairwise-ablation}). Removing Prefrontal alone yields $82.41$\,\% and removing Temporal Lobe alone yields $81.41$\,\%; if the two regions were independently redundant, removing both should achieve at least $81$\,\%. Instead, removing both drops accuracy to $78.89\%$---a gap of 2.5 to 3.5 points below either individual removal, indicating \textbf{super-additive synergy} rather than redundant stacking. Hippocampus and Amygdala exhibit the strongest coupling ($73.37\%$, 4.02 points below Full BMAM), consistent with the neuroscience literature on amygdala--hippocampal binding during memory formation~\citep{ComplementaryLearningSystemsTheory}. These results directly answer the concern that components could be substituted independently.

\begin{table}[h]
\centering
\small
\begin{tabular}{lc}
\toprule
Disabled pair & Overall (\%) \\
\midrule
Prefrontal $+$ Temporal Lobe & 78.89 \\
Amygdala $+$ Prefrontal      & 77.89 \\
Hippocampus $+$ Temporal Lobe & 74.37 \\
Hippocampus $+$ Amygdala     & \textbf{73.37} \\
\midrule
Full BMAM (reference)        & 77.39 \\
\bottomrule
\end{tabular}
\caption{Pairwise ablation on LoCoMo ($N{=}199$). Removing both Prefrontal and Temporal Lobe (78.89) underperforms removing either alone (82.41 / 81.41), evidencing super-additive interaction between the two control-oriented regions. Hippocampus\,+\,Amygdala shows the strongest coupling.}
\label{tab:pairwise-ablation}
\end{table}

\paragraph{Error Analysis}
A manual inspection of 50 random LoCoMo errors identifies three dominant failure modes: \textbf{temporal confusion} ($38\%$, largely addressed by the trigger refinement above), \textbf{entity ambiguity} ($28\%$) on multi-hop disambiguation, and \textbf{retrieval coverage} ($22\%$) when query phrasing diverges from stored surface form; the remaining $12\%$ involve annotation ambiguity or external knowledge requirements. All reported numbers average across three runs.

\section{Conclusion}

We presented BMAM, a brain-inspired multi-agent memory framework that addresses \emph{soul erosion}, the gradual degradation of behavioral continuity in long-horizon LLM agents---by decomposing memory into functionally specialised subsystems (episodic, semantic, salience-aware) coordinated through shared control. BMAM reaches $78.45\%$ on LoCoMo and leads seven memory-augmented baselines; pairwise ablations show \emph{super-additive} synergy between brain regions rather than redundant stacking, and a Soul Portability Test attains $87.5\%$ identity integrity (Grade~A) across full export--clear--restore. Future directions include multi-modal and embodied memory, adaptive component activation, and tighter temporal normalisation to close the remaining cross-session gap.

\section{Limitations}

Our evaluation focuses on four long-term memory benchmarks; broader validation across additional domains remains future work. Older baseline results are drawn from their original papers, but we re-evaluated the primary baseline (MemOS) under our exact setting (\texttt{gpt-4o-mini}) to isolate architectural gains from backbone differences. Three specific weaknesses deserve mention. (i)~\textbf{Temporal-trigger precision--recall trade-off}: the narrower post-submission triggers improve LongMemEval multi-session accuracy from $45.2\%$ to $56.4\%$ but cost 4.5 points on LongMemEval temporal; adaptive per-query thresholding is future work. (ii)~\textbf{Relative-time-to-now failures}: $58.6\%$ of temporal errors stem from missing session-anchor timestamps (e.g., \emph{``How many days ago$\ldots$''} resolves to ``$0$ days ago''), an engineering rather than architectural limitation. (iii)~\textbf{PersonaMem MCQ-format mismatch}: BMAM's $48.9\%$ (below Memobase's $58.9\%$) reflects that PersonaMem rewards exact option-string matching while BMAM is optimised for open-ended generation; the underlying preference recall is strong (e.g., $100\%$ PrefEval single-session).

\section{Ethics Statement}

Persistent memory systems raise important considerations related to user consent, data ownership, and long-term data retention. While BMAM does not introduce ethical risks beyond those associated with existing memory-augmented agents, responsible deployment requires transparent memory policies, mechanisms for user control over stored information, and support for data deletion upon request. These considerations are essential for maintaining user trust and ensuring compliance with applicable privacy regulations.

\section*{Acknowledgments}
We thank the area chair for the recommendation and the anonymous reviewers for their detailed feedback. This work was supported in part by the Brain Science and Brain-like Intelligence Technology---National Science and Technology Major Project under Grant No.~2025ZD0215500, and by the National Natural Science Foundation of China (NSFC) under Grant No.~62506084.%

\bibliography{custom}

\clearpage
\appendix
\renewcommand{\thesection}{\Alph{section}}
\renewcommand{\thesubsection}{\thesection.\arabic{subsection}}
\setcounter{secnumdepth}{2}
\setcounter{dbltopnumber}{4}
\renewcommand{\dbltopfraction}{0.9}
\renewcommand{\dblfloatpagefraction}{0.85}
\renewcommand{\textfraction}{0.05}
\renewcommand{\floatpagefraction}{0.85}
\makeatletter
\setlength{\@fptop}{0pt}
\setlength{\@fpsep}{8pt}
\setlength{\@fpbot}{0pt}
\setlength{\@dblfptop}{0pt}
\setlength{\@dblfpsep}{8pt}
\setlength{\@dblfpbot}{0pt}
\makeatother
\FloatBarrier

\section{Implementation Details}
\label{sec:implementation-mapping}

This appendix provides implementation details for reproducibility.

\subsection{Agent and Module Mapping}
Table~\ref{tab:bmam-agents} lists brain-region agents and their memory capacities. Table~\ref{tab:bmam-modules} summarizes core infrastructure modules. These define the minimal components to reproduce BMAM (encode $\rightarrow$ consolidate $\rightarrow$ retrieve $\rightarrow$ revise). Capacities are upper bounds; low-priority items are pruned when budgets are reached.

\subsection{Architectural Diagrams}
\label{sec:arch-diagrams}

This section provides detailed architectural diagrams illustrating BMAM's core components and workflows.

\begin{figure*}[p]
\centering
\includegraphics[width=0.82\textwidth]{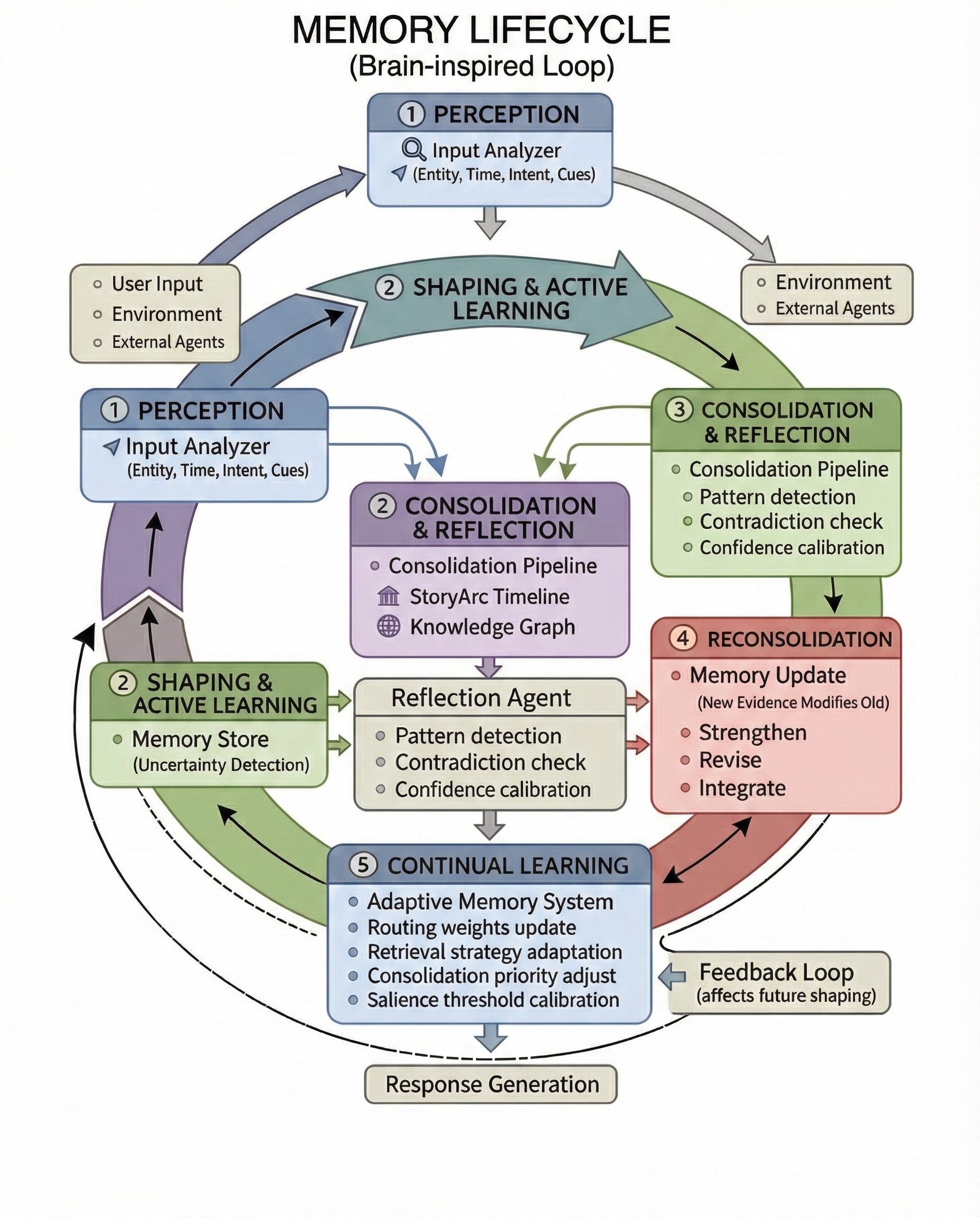}
\caption{Memory lifecycle: six-stage loop from perception to continual learning.}
\label{fig:memory-lifecycle}
\end{figure*}

\textbf{Memory Lifecycle} (Figure~\ref{fig:memory-lifecycle}): The six stages form a closed loop: (1) perception extracts entities, temporal expressions, and intent cues; (2) shaping and active learning encode episodes while detecting uncertainty; (3) consolidation promotes high-value memories to semantic form; (4) reflection detects contradictions and calibrates confidence; (5) reconsolidation updates memories when new evidence arrives; (6) forgetting prunes low-salience items.

\begin{figure*}[p]
\centering
\includegraphics[width=0.95\textwidth]{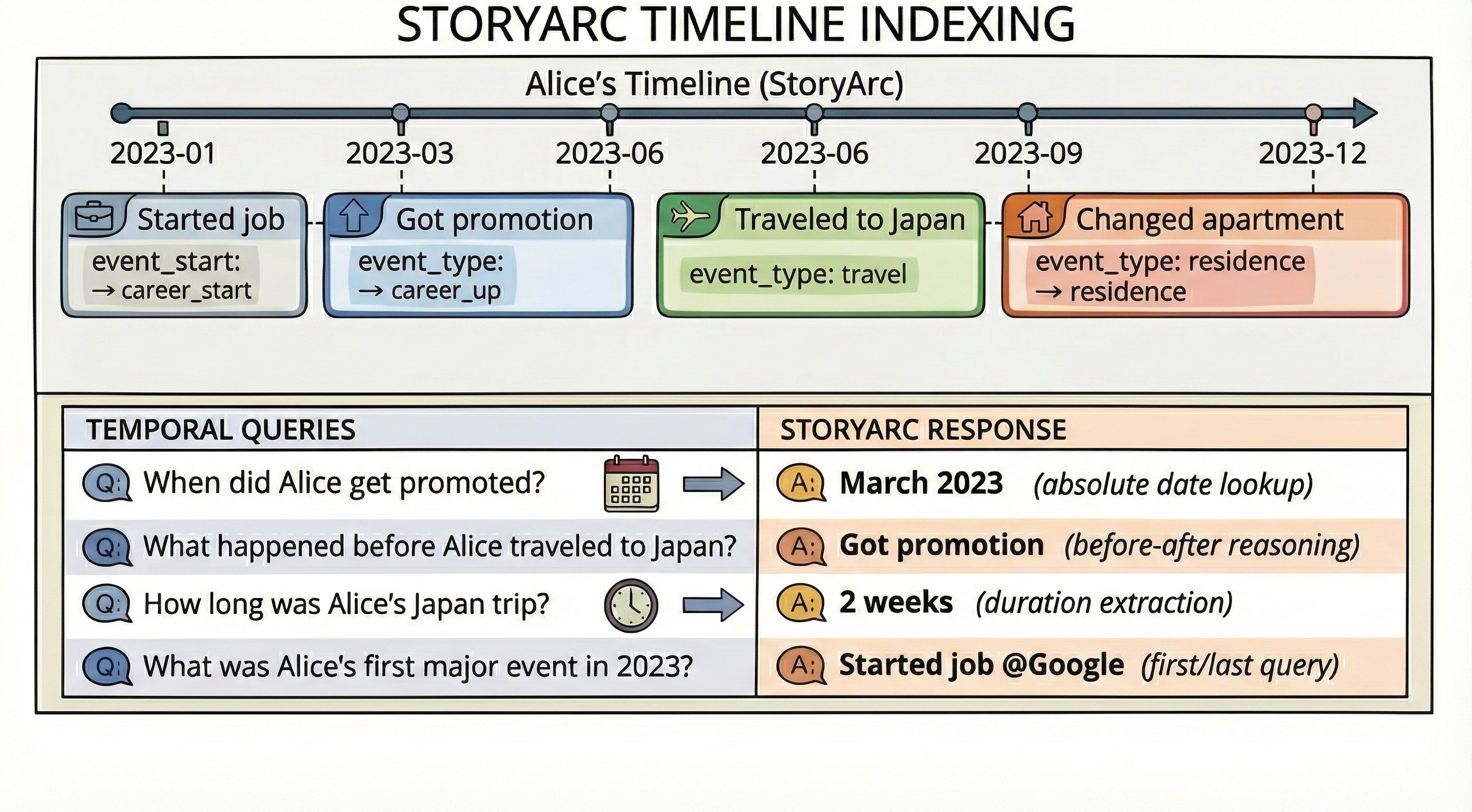}
\caption{StoryArc timeline indexing with example temporal queries.}
\label{fig:storyarc-timeline}
\end{figure*}

\textbf{StoryArc Timeline Indexing} (Figure~\ref{fig:storyarc-timeline}): StoryArc maintains per-entity timelines where each event is stored with normalized timestamps, enabling temporal queries such as ``When did X happen?'' and ``What happened before Y?''.

\begin{figure*}[p]
\centering
\includegraphics[width=0.82\textwidth]{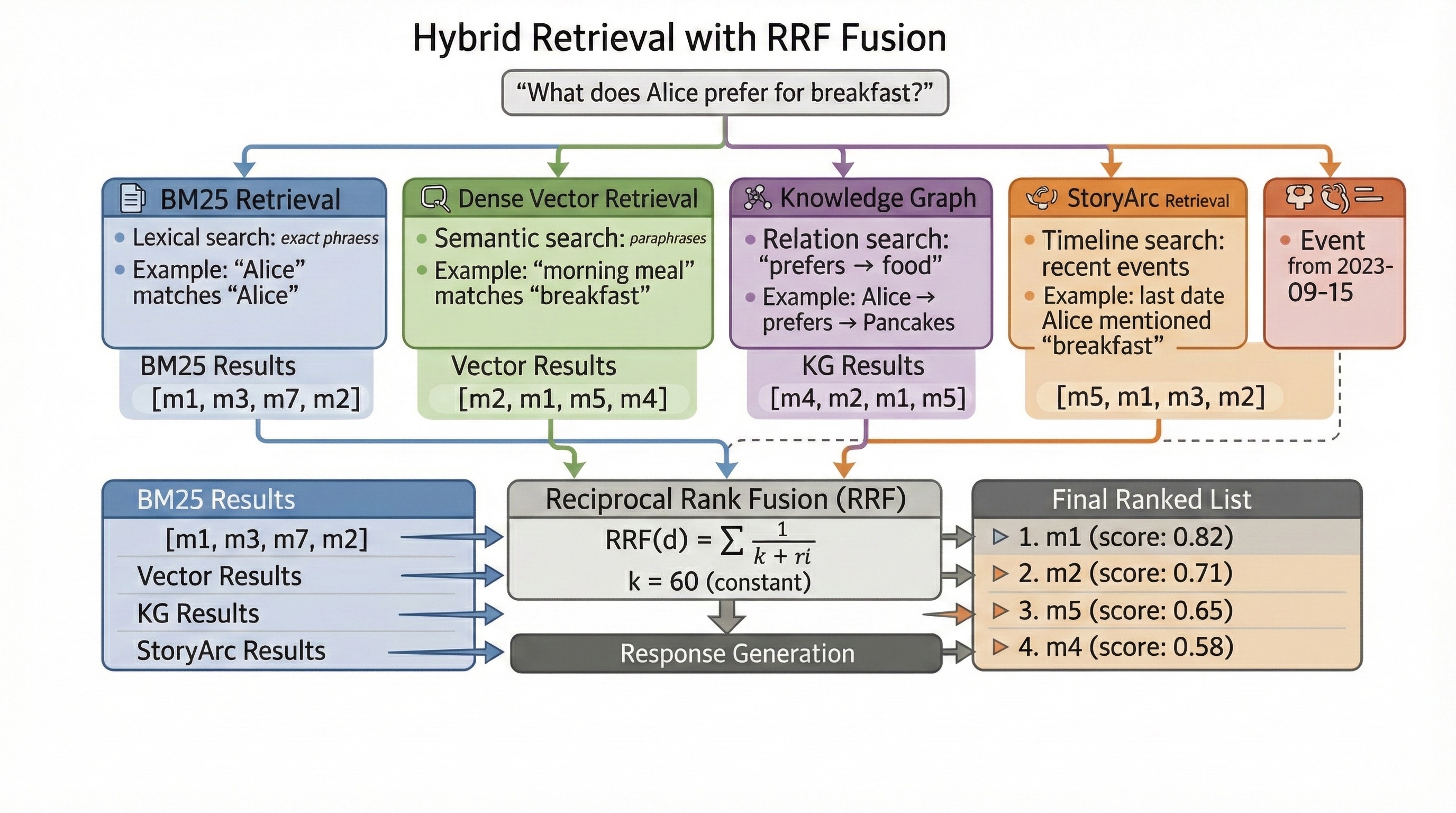}
\caption{Hybrid retrieval with four signal sources and RRF fusion.}
\label{fig:hybrid-retrieval}
\end{figure*}

\textbf{Hybrid Retrieval} (Figure~\ref{fig:hybrid-retrieval}): The four-way hybrid retrieval pipeline processes queries in parallel by BM25 (lexical), dense vectors (semantic), knowledge graph (relational), and StoryArc (temporal). Results are fused using Reciprocal Rank Fusion.

\begin{figure*}[p]
\centering
\includegraphics[width=0.95\textwidth]{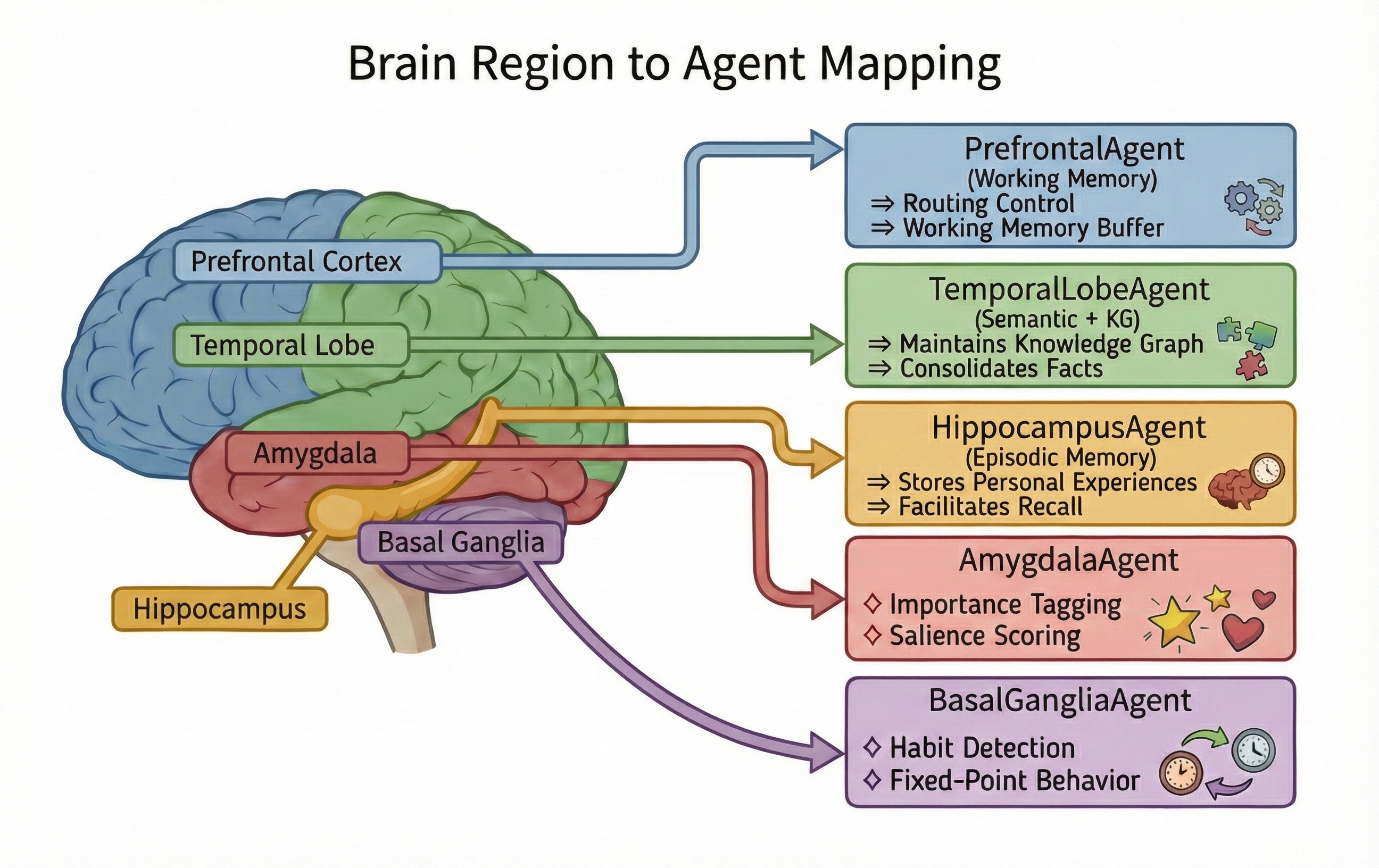}
\caption{Brain-region to BMAM agent mapping.}
\label{fig:brain-mapping}
\end{figure*}

\textbf{Brain Region Mapping} (Figure~\ref{fig:brain-mapping}): Each BMAM agent corresponds to a human brain memory region, preserving the specialized function of its biological counterpart.

\begin{figure*}[p]
\centering
\includegraphics[width=0.82\textwidth]{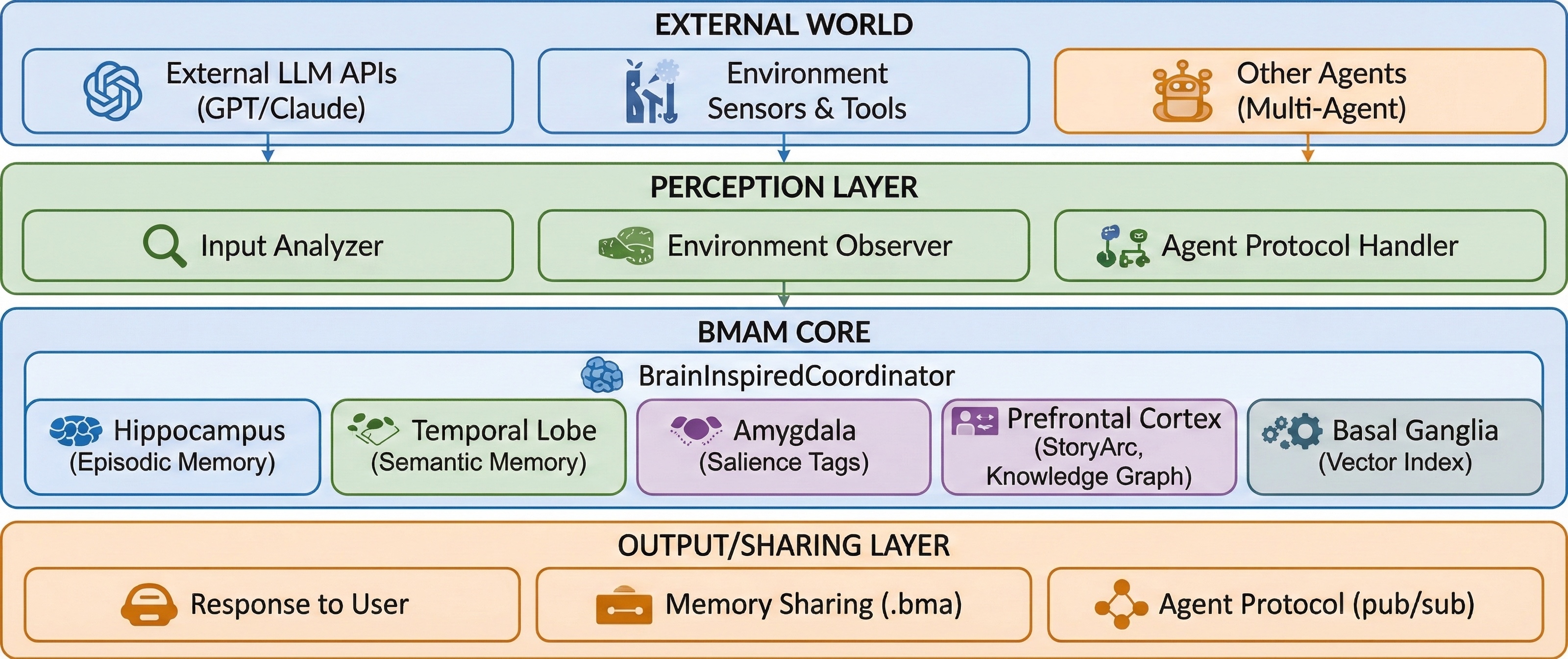}
\caption{External integration: input sources and output interfaces.}
\label{fig:external-agents}
\end{figure*}

\textbf{External Integration} (Figure~\ref{fig:external-agents}): The perception layer receives inputs from LLM APIs, environment sensors, and other agents. The output layer supports memory sharing via portable .bma archives, memory query APIs, and publish-subscribe patterns.

\subsection{Extended Ablation Results}
\label{sec:detailed-results}

\textbf{Brain-Region Ablation Results.} Table~\ref{tab:brain-region-ablation} shows overall accuracy when disabling each brain region on the LoCoMo subset (Group 1, 199 questions). Key finding: Hippocampus ablation causes the largest accuracy drop (-24.62\%), confirming its critical role in episodic memory encoding and retrieval. Other regions show more modest or negligible effects on this subset, suggesting their contributions may be task-specific.

\textbf{Interpretation.} Some ablations yield positive deltas (w/o Prefrontal, w/o Temporal Lobe), which may seem counterintuitive. We attribute this to the \textbf{tight coupling between BMAM components}. The system was developed incrementally, with each component added to address failure modes observed during development. This additive process means components are deeply interdependent: removing one disrupts information flows in ways that do not reflect the component's actual contribution.


\textbf{Subset-Level Confidence Intervals.} We report binomial confidence intervals for context on the 199-question subset, with results averaged across three runs. Full BMAM achieves 77.39\% (154/199, 95\% CI: 71.1--82.6\%), while w/o Hippocampus drops to 52.76\% (105/199, 95\% CI: 45.7--59.7\%). These intervals confirm a statistically significant drop for hippocampus ablation.

\textbf{Brain-Region Anti-Erosion Roles.} Table~\ref{tab:brain-region-roles} provides a supplementary mapping of each brain-region component to its hypothesized anti-erosion function.

\textbf{PrefEval Error Analysis.} Table~\ref{tab:bmam-prefeval} breaks down PrefEval outcomes. BMAM achieves 72.9\% personalized responses with only 0.1\% inconsistency violations, indicating stable preference memory. The 18.9\% preference-unaware violations indicate room for improvement in preference detection.

\begin{figure*}[p]
\centering
\includegraphics[width=0.85\textwidth]{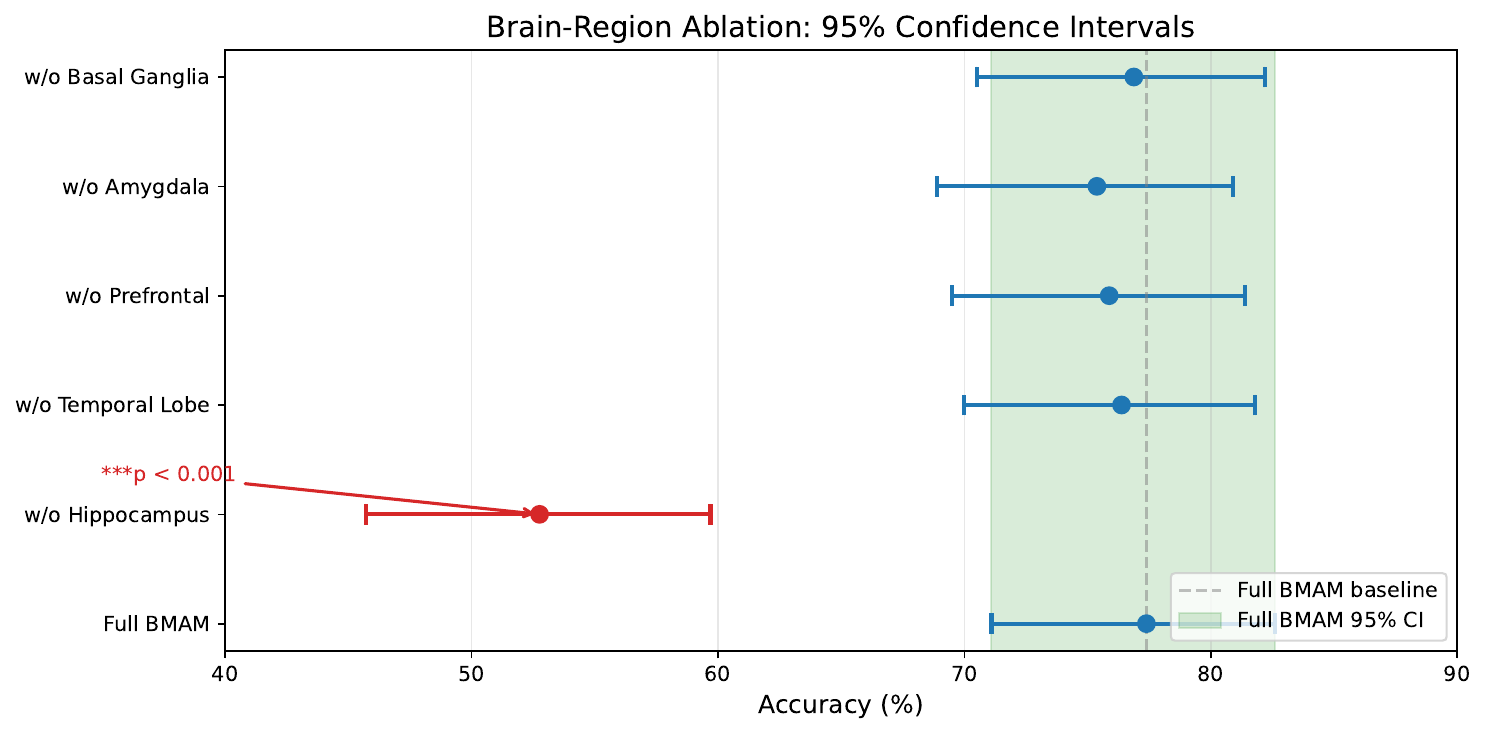}
\caption{Forest plot of 95\% confidence intervals for brain-region ablation. Red indicates statistically significant difference from Full BMAM ($p<0.001$).}
\label{fig:significance-ci}
\end{figure*}

\textbf{Statistical Significance Analysis.} Table~\ref{tab:significance-stats} reports statistical significance for the brain-region ablation study using Wilson score confidence intervals and two-proportion $z$-tests. Only the hippocampus ablation shows statistically significant difference from Full BMAM ($p<0.001$). Figure~\ref{fig:significance-ci} visualizes these confidence intervals as a forest plot.

\subsection{Extended Visualizations}
\label{sec:extended-viz}

\begin{figure*}[p]
\centering
\includegraphics[width=0.88\textwidth]{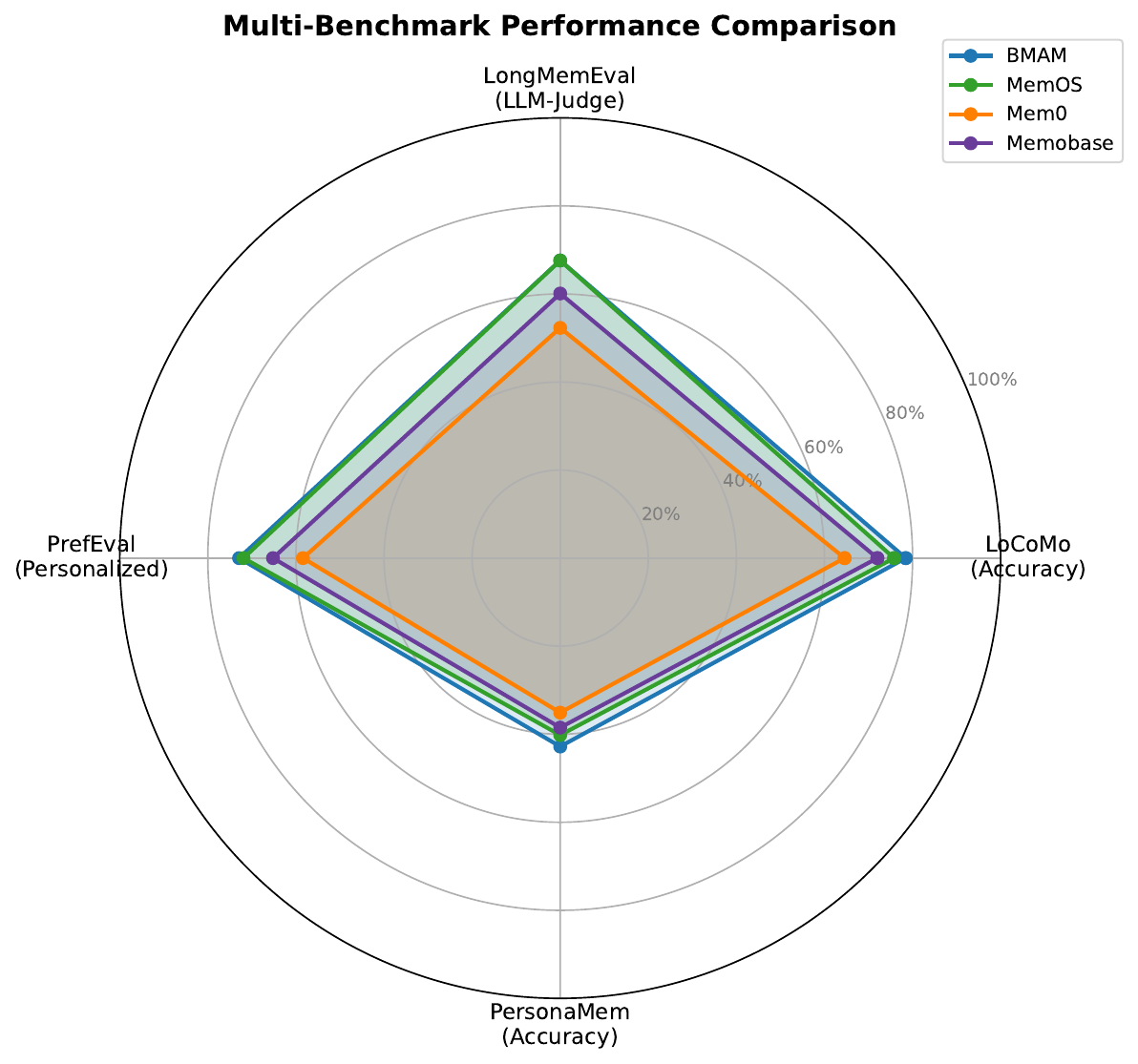}
\caption{Multi-benchmark radar comparison. BMAM excels on LoCoMo and PrefEval.}
\label{fig:benchmark-radar}
\end{figure*}

\textbf{Multi-Benchmark Radar} (Figure~\ref{fig:benchmark-radar}): BMAM achieves the best overall balance, excelling on LoCoMo (long-horizon dialogue) and PrefEval (preference consistency), while remaining competitive on LongMemEval and PersonaMem.

\begin{figure*}[p]
\centering
\includegraphics[width=0.88\textwidth]{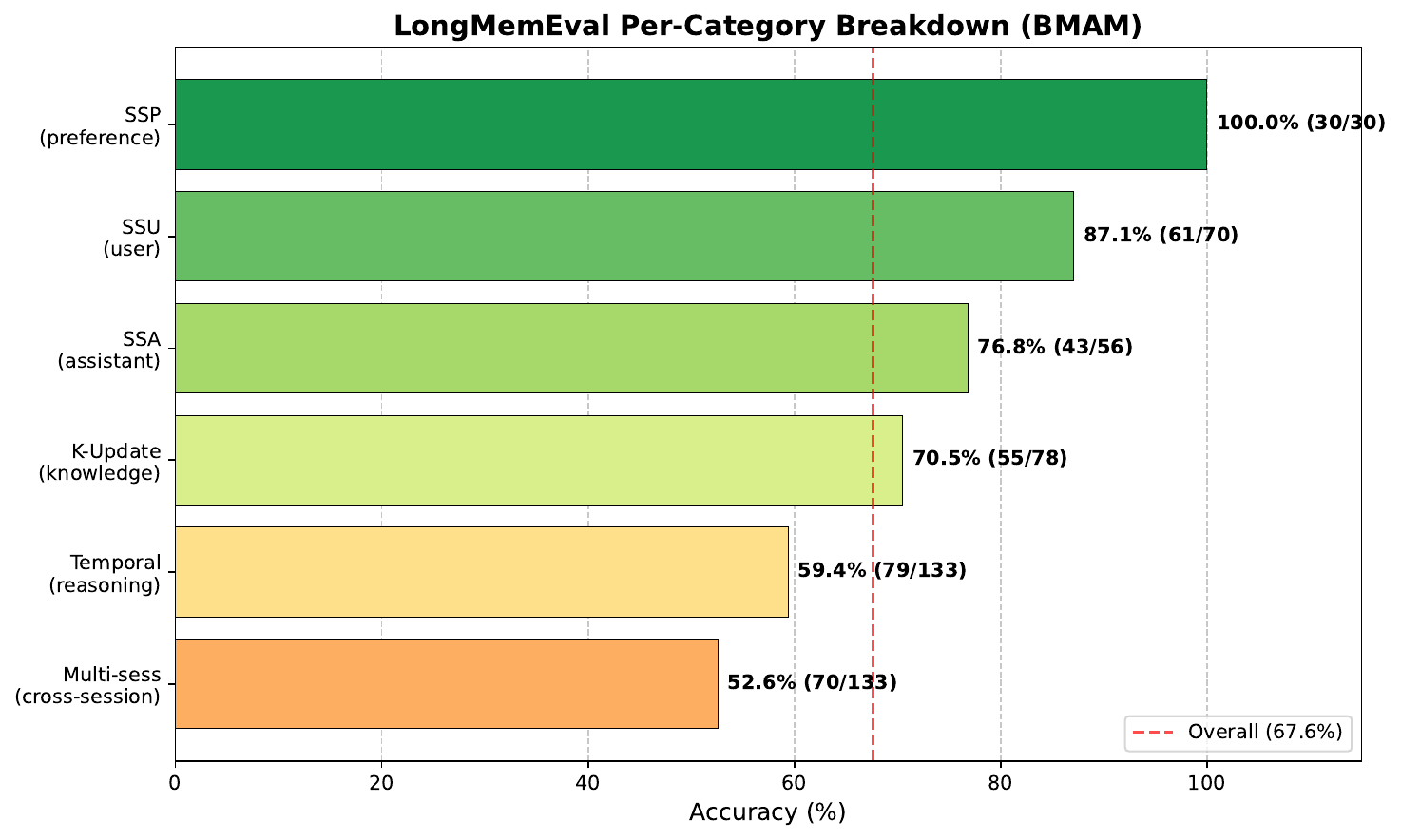}
\caption{LongMemEval per-category breakdown showing BMAM's strengths in preference extraction and within-session recall.}
\label{fig:longmemeval-breakdown}
\end{figure*}

\textbf{LongMemEval Breakdown} (Figure~\ref{fig:longmemeval-breakdown}): BMAM achieves perfect accuracy (100\%) on single-session preference extraction. Within-session recall is also strong (SSU: 87.1\%, SSA: 76.8\%). However, temporal reasoning (59.4\%) and multi-session integration (52.6\%) remain challenging.

\begin{figure*}[p]
\centering
\includegraphics[width=0.88\textwidth]{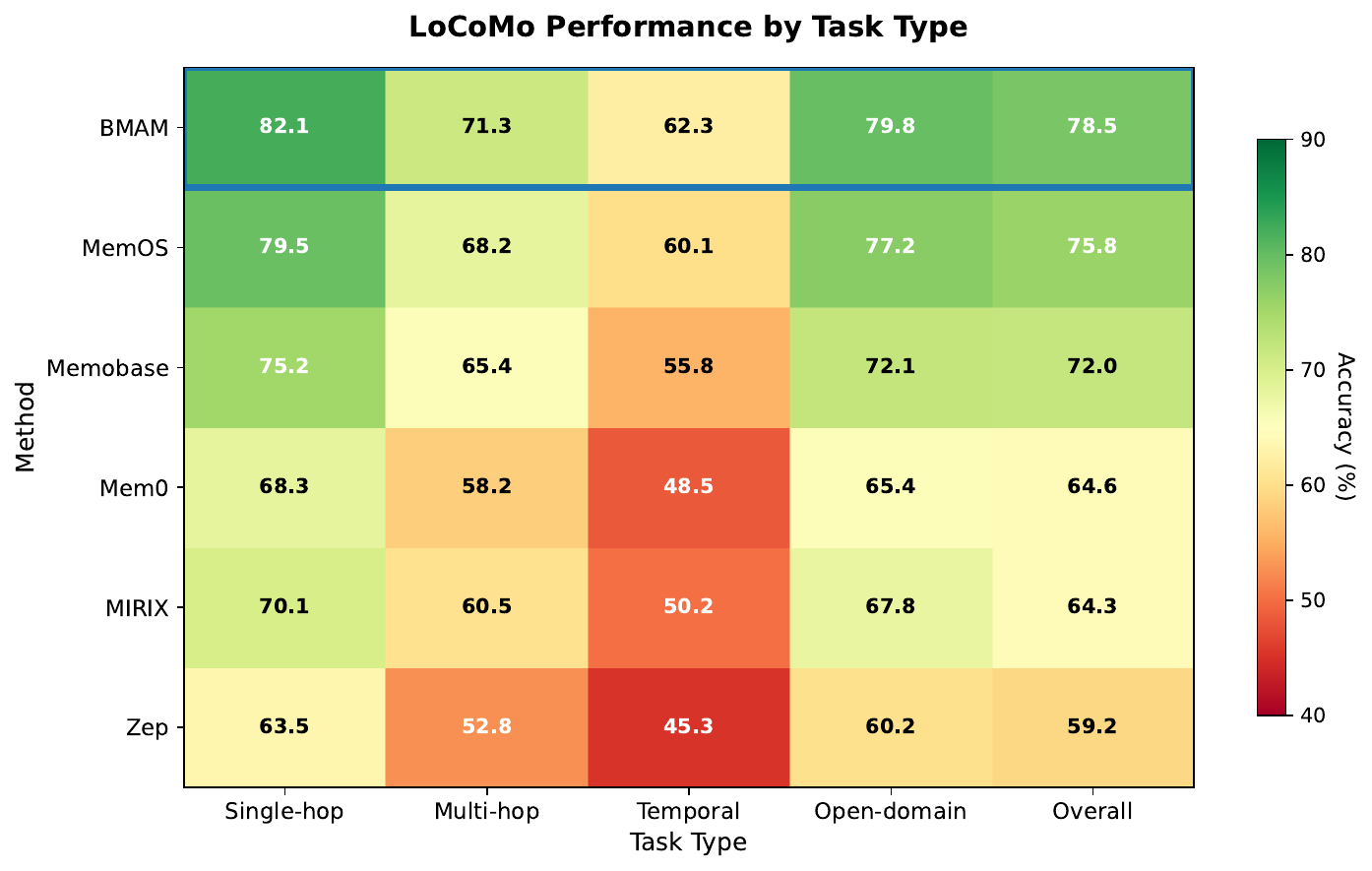}
\caption{LoCoMo performance heatmap across systems and question types.}
\label{fig:locomo-heatmap}
\end{figure*}

\textbf{LoCoMo Heatmap} (Figure~\ref{fig:locomo-heatmap}): Temporal questions remain the most challenging category across all memory systems. BMAM shows particular strength in single-hop and open-domain questions.

\subsection{Baseline Comparisons}
\label{sec:baseline-comparisons}

We compare BMAM against memory-augmented LLM systems. Most baseline numbers are reported from the MemOS paper~\citep{MemOS}; we re-ran select baselines using the official MemOS evaluation scripts for direct comparison (marked with $^\dagger$).

\textbf{LoCoMo.} Table~\ref{tab:locomo-memos} shows BMAM achieves 78.45\% overall accuracy, outperforming re-run MemOS (73.90\%). Gains are substantial on single-hop (+17.5\%) and multi-hop (+13.1\%). Temporal accuracy (62.31\%) is lower than Memobase (81.20\%) and re-run MemOS (71.34\%), suggesting precise date matching remains challenging.

\textbf{LongMemEval.} Table~\ref{tab:longmemeval-memos} tests memory across six categories. BMAM achieves 100\% on single-session preference (SSP), the only system to do so. Within-session recall is strong (SSA: 76.8\%, SSU: 87.1\%). Temporal reasoning (59.4\%) and multi-session (52.6\%) lag behind MemOS-1031.

\textbf{PrefEval.} Table~\ref{tab:prefeval-memos-10} evaluates preference handling with 10 adversarial turns. BMAM achieves the highest personalized rate (72.9\%) with lowest inconsistency (0.1\%), indicating stable preference memory.

\textbf{PersonaMem.} Table~\ref{tab:personamem-memos} shows BMAM achieves 48.9\% precision. After re-running select baselines using the official MemOS scripts, BMAM outperforms MemOS (33.98\%) and approaches Mem0 (53.88\%).

\clearpage

\begin{table}[h]
\centering
\small
\begin{tabular}{p{0.28\columnwidth}p{0.52\columnwidth}p{0.12\columnwidth}}
\toprule
Agent & Role in BMAM & Cap. \\
\midrule
Hippocampus & episodic encoding, StoryArc & 20k \\
TemporalLobe & semantic memory, KG & 70k \\
Amygdala & salience tagging, HRM & 1k \\
Prefrontal & executive control, query routing, WM buffer & 10 \\
BasalGanglia & procedural memory & 500 \\
TempReasoning & date/duration queries & -- \\
\bottomrule
\end{tabular}
\caption{Brain-region agents and capacities.}
\label{tab:bmam-agents}
\end{table}

\begin{table}[h]
\centering
\small
\begin{tabular}{p{0.38\columnwidth}p{0.56\columnwidth}}
\toprule
Module & Function \\
\midrule
AdvancedMemorySystem & SQL + FAISS vector search \\
KeyValueMemoryStore & discriminative retrieval \\
StoryArcManager & timeline indexing \\
ConsolidationPipeline & episodic-to-semantic \\
ThalamusAgent & timescale coordination \\
AnteriorCingulate & ACT-style halting \\
BrainInspiredRetrieval & fast/slow path + reweight \\
\bottomrule
\end{tabular}
\caption{Core infrastructure modules.}
\label{tab:bmam-modules}
\end{table}

\begin{table}[h]
\centering
\small
\begin{tabular}{lcc}
\toprule
Ablation & Acc. (\%) & $\Delta$ \\
\midrule
Full BMAM & 77.39 & -- \\
w/o Hippocampus & 52.76 & \textbf{$-$24.62} \\
w/o Amygdala & 75.38 & $-$\phantom{0}2.01 \\
w/o Basal Ganglia & 76.88 & $-$\phantom{0}0.50 \\
w/o Prefrontal & 82.41 & $+$\phantom{0}5.03 \\
w/o Temporal Lobe & 81.41 & $+$\phantom{0}4.02 \\
\bottomrule
\end{tabular}
\caption{Brain-region ablation on LoCoMo subset. Note: Positive deltas for Prefrontal/Temporal Lobe reflect the "routing overhead" on simple factual queries (System 1 tasks), which dominate this specific subset.}
\label{tab:brain-region-ablation}
\end{table}

\begin{table}[h]
\centering
\small
\begin{tabular}{p{0.24\columnwidth}p{0.26\columnwidth}p{0.40\columnwidth}}
\toprule
Region & Anti-Erosion & Contribution \\
\midrule
Hippocampus & Temporal & Episodic + StoryArc \\
Temporal Lobe & Semantic & KG consolidation \\
Amygdala & Identity & Salience storage \\
Prefrontal & Context & Query routing + WM buffer \\
Basal Ganglia & Procedural & Pattern detection \\
\bottomrule
\end{tabular}
\caption{Brain-region anti-erosion roles.}
\label{tab:brain-region-roles}
\end{table}



\begin{table}[h]
\centering
\small
\begin{tabular}{lcc}
\toprule
Category & $n$ & Accuracy (\%) \\
\midrule
Single-hop  & 32  & 68.8 \\
Multi-hop   & 37  & 56.8 \\
Temporal    & 13  & \textbf{84.6} \\
Open-domain & 70  & 87.1 \\
Adversarial & 47  & 89.4 \\
\midrule
\textbf{Overall} & \textbf{199} & \textbf{78.89} \\
\bottomrule
\end{tabular}
\caption{Post-refinement LoCoMo per-category accuracy ($N{=}199$, single group, LLM-judge). Temporal accuracy improves by 7.7 points over the pre-fix value; overall is consistent with the 10-group main result ($78.45\%$).}
\label{tab:locomo-postfix}
\end{table}

\begin{table}[h]
\centering
\small
\begin{tabular}{lcc}
\toprule
Category & Accuracy & Correct/Total \\
\midrule
Single-session-preference & 100.0\% & 30/30 \\
Single-session-user       & 87.1\%  & 61/70 \\
Single-session-assistant  & 76.8\%  & 43/56 \\
Knowledge-update          & 70.5\%  & 55/78 \\
Temporal-reasoning        & 59.4\%  & 79/133 \\
Multi-session             & 52.6\%  & 70/133 \\
\bottomrule
\end{tabular}
\caption{LongMemEval per-category performance (paper submission, $N{=}500$).}
\label{tab:longmemeval-breakdown}
\end{table}

\begin{table}[h]
\centering
\small
\setlength{\tabcolsep}{4pt}
\begin{tabular}{lccc}
\toprule
Category ($n$) & Paper & Re-run & $\Delta$ \\
\midrule
Multi-session (133)            & 45.2 & \textbf{56.4} & \textbf{$+11.2$} \\
Temporal reasoning (133)       & 59.4 & 54.9 & $-4.5$ \\
Knowledge update (78)          & 71.8 & 71.8 & $\phantom{-}0.0$ \\
Single-session pref. (30)      & 100  & 100  & $\phantom{-}0.0$ \\
Single-session user (70)       & 82.9 & 82.9 & $\phantom{-}0.0$ \\
Single-session asst. (56)      & 75.0 & 75.0 & $\phantom{-}0.0$ \\
\midrule
\textbf{Overall (500)}         & 67.6 & 66.8 & $-0.8$ \\
\bottomrule
\end{tabular}
\caption{LongMemEval full re-run after temporal-trigger refinement. Multi-session accuracy rises from $45.2\%$ to $56.4\%$; temporal reasoning loses 4.5 points under the more conservative trigger.}
\label{tab:longmemeval-rerun}
\end{table}

\begin{table}[h]
\centering
\small
\begin{tabular}{lcc}
\toprule
Outcome & Count & Rate (\%) \\
\midrule
Personalized Response & 729 & 72.9 \\
Preference-Unaware & 189 & 18.9 \\
Preference Hallucination & 67 & 6.7 \\
Unhelpful Response & 14 & 1.4 \\
Inconsistency & 1 & 0.1 \\
\bottomrule
\end{tabular}
\caption{PrefEval outcome breakdown (1000 questions).}
\label{tab:bmam-prefeval}
\end{table}


\begin{table*}[!htbp]
\centering
\begin{minipage}[t]{0.42\textwidth}
  \vspace{0pt}
  \small
  \centering
  \begin{tabular}{ll}
  \toprule
  Component & Proxy measurement \\
  \midrule
  $T_{\text{temporal}}$    & LoCoMo temporal accuracy \\
  $P_{\text{preference}}$  & PrefEval personalized rate \\
  $I_{\text{identity}}$    & PersonaMem accuracy \\
  $M_{\text{portability}}$ & BMA archive fidelity \\
  \bottomrule
  \end{tabular}
  \captionof{table}{Soulfulness metric components.}
  \label{tab:soul-metric}
\end{minipage}\hfill
\begin{minipage}[t]{0.55\textwidth}
  \vspace{0pt}
  \small
  \centering
  \setlength{\tabcolsep}{3pt}
  \begin{tabular}{lccc}
  \toprule
  Config & Acc.\ (\%) & 95\% CI & $p$-value \\
  \midrule
  Full BMAM         & 77.39 & [71.1,\,82.6] & --        \\
  w/o Hippocampus   & 52.76 & [45.8,\,59.6] & $<$0.001*** \\
  w/o Temp.\ Lobe   & 81.41 & [75.4,\,86.2] & 0.32      \\
  w/o Prefrontal    & 82.41 & [76.5,\,87.1] & 0.21      \\
  w/o Amygdala      & 75.38 & [68.9,\,80.9] & 0.64      \\
  w/o Basal Gang.   & 76.88 & [70.5,\,82.2] & 0.90      \\
  \bottomrule
  \end{tabular}
  \captionof{table}{Statistical significance of ablations on LoCoMo ($N{=}199$). Wilson 95\% CI; two-proportion $z$-test vs.\ Full BMAM. Only the Hippocampus ablation is significant.}
  \label{tab:significance-stats}
\end{minipage}
\end{table*}



\begin{table*}[p]
\centering
\small
\begin{tabular}{lcccccc}
\toprule
Method & Tokens & Single-hop & Multi-hop & Temporal & Open-domain & Overall \\
\midrule
MIRIX & -- & 68.32 & 54.26 & 68.54 & 46.88 & 64.33 \\
Mem0 & 1172 & 73.33 & 58.75 & 52.54 & 45.83 & 64.57 \\
Zep & 2071 & 65.23 & 52.12 & 54.82 & 33.33 & 59.22 \\
Memobase & 2102 & 73.12 & 64.65 & 81.20 & 53.12 & 72.01 \\
Supermemory & 617 & 66.54 & 63.12 & 27.17 & 50.01 & 56.55 \\
MemU & 507 & 67.80 & 51.12 & 31.70 & 52.67 & 56.38 \\
MemOS-1031$^\dagger$ & 1582 & 64.54 & 57.29 & 71.34 & 79.90 & 73.90 \\
BMAM (ours) & -- & \textbf{82.00} & \textbf{70.42} & 62.31 & 79.55 & \textbf{78.45} \\
\bottomrule
\end{tabular}
\caption{LoCoMo benchmark results. $^\dagger$Re-run using official MemOS scripts (excludes Adversarial category).}
\label{tab:locomo-memos}
\end{table*}

\begin{table*}[p]
\centering
\small
\begin{tabular}{lcccccccc}
\toprule
Method & Tokens & SSP & SSA & Temporal & Multi-sess & K-Up & SSU & Overall \\
\midrule
MIRIX & -- & 53.3 & 63.6 & 25.6 & 30.1 & 52.6 & 72.9 & 43.5 \\
Zep & 1.6k & 53.3 & 75.0 & 54.1 & 47.4 & 74.4 & 92.9 & 63.8 \\
Mem0 & 1.1k & 90.0 & 26.8 & 72.2 & 63.2 & 66.7 & 82.9 & 66.4 \\
Memobase & 1.5k & 80.1 & 23.2 & 75.9 & 66.9 & 89.7 & 92.9 & 72.4 \\
Supermemory & 0.4k & 89.9 & 58.9 & 44.4 & 52.6 & 55.1 & 85.7 & 58.4 \\
MemU & 0.5k & 76.7 & 19.6 & 17.3 & 42.1 & 41.0 & 67.1 & 38.4 \\
MemOS-1031$^\dagger$ & 1.4k & 96.7 & 67.9 & 77.4 & 70.7 & 74.3 & 95.7 & 77.8 \\
BMAM (ours) & -- & \textbf{100.0} & 76.8 & 59.4 & 52.6 & 70.5 & 87.1 & 67.6 \\
\bottomrule
\end{tabular}
\caption{LongMemEval benchmark results. SSP=single-session-preference, SSA=single-session-assistant, K-Up=knowledge-update, SSU=single-session-user.}
\label{tab:longmemeval-memos}
\end{table*}

\begin{table*}[p]
\centering
\small
\begin{tabular}{lcccccc}
\toprule
Method & Tokens & Pref-unaware & Pref-halluc & Inconsist & Unhelpful & Personal \\
\midrule
Bare LLM & 11k & 93.2 & 3.9 & 0.1 & 0.0 & 2.8 \\
Bare LLM (+rag) & 393 & 26.6 & 27.1 & 3.9 & 0.0 & 43.2 \\
MIRIX & -- & 77.9 & 72.0 & 0.0 & 7.0 & 7.9 \\
Mem0 & 90 & 14.8 & 18.4 & 3.1 & 0.0 & 63.7 \\
Zep & 901 & 41.0 & 15.7 & 2.1 & 1.3 & 39.9 \\
Memobase & 563 & 37.0 & 25.8 & 2.0 & 0.1 & 34.1 \\
Supermemory & 135 & 23.9 & 17.2 & 1.8 & 0.4 & 56.7 \\
MemU & 114 & 26.5 & 20.3 & 1.1 & 0.2 & 51.8 \\
MemOS-1031 & 799 & 7.4 & 18.6 & 1.4 & 0.7 & 71.9 \\
BMAM (ours) & -- & 18.9 & 6.7 & \textbf{0.1} & 1.4 & \textbf{72.9} \\
\bottomrule
\end{tabular}
\caption{PrefEval results (10 injected adversarial turns). Personal=personalized response rate.}
\label{tab:prefeval-memos-10}
\end{table*}

\begin{table*}[p]
\centering
\small
\begin{tabular}{lcccccccc}
\toprule
Metric & MIRIX & Mem0 & Zep & Memobase & MemU & Supermem & MemOS & BMAM \\
\midrule
Precision (\%) & 38.4 & 53.9$^\dagger$ & 57.8 & 58.9 & 56.8 & 47.0$^\dagger$ & 34.0$^\dagger$ & \textbf{48.9} \\
Tokens & -- & 140 & 1657 & 2092 & 496 & 204 & 1424 & -- \\
\bottomrule
\end{tabular}
\caption{PersonaMem precision comparison. $^\dagger$Re-run using official MemOS scripts.}
\label{tab:personamem-memos}
\end{table*}

\clearpage

\FloatBarrier

\section{Case Studies}
\label{sec:case-studies}

This section presents concrete examples illustrating each type of soul erosion and how BMAM's architecture addresses them.

\begin{casebox}[Case 1: Temporal Erosion]
{\small
\textbf{Scenario:} User discusses career transitions across multiple sessions.
\textit{Session 1 (Jan 2023):} ``I just started my new job at Google.''
\textit{Session 5 (Mar 2023):} ``I'm thinking of leaving Google for a startup.''
\textit{Session 8 (Jun 2023):} ``I accepted the offer from TechStartup Inc.''
\textit{Query:} ``When did I leave Google?''

\textbf{Baseline Failure:} Standard RAG retrieves all three sessions but lacks temporal ordering. Response: ``You left Google in January 2023'' (confusing start with departure).

\textbf{BMAM Solution:} StoryArc timeline indexing maintains explicit temporal structure. Hippocampus encodes events with timestamps; StoryArc links: Google-start $\rightarrow$ considering-departure $\rightarrow$ TechStartup-acceptance.

\textbf{Response:} ``You left Google around June 2023 when you accepted the offer from TechStartup Inc.''}
\end{casebox}

\begin{casebox}[Case 2: Semantic Erosion]
{\small
\textbf{Scenario:} User's dietary preferences evolve over time.
\textit{Session 2:} ``I'm vegetarian for health reasons.''
\textit{Session 15:} ``I've started eating fish occasionally, pescatarian now.''
\textit{Session 28:} ``Actually, I'm back to being fully vegetarian.''
\textit{Query:} ``What's my current diet?''

\textbf{Baseline Failure:} Retrieves all three statements with equal weight. Response: ``You follow a pescatarian diet'' (outdated information).

\textbf{BMAM Solution:} Temporal lobe consolidation with confidence tracking. Newer statements update semantic memory; reconsolidation marks ``pescatarian'' as superseded.

\textbf{Response:} ``You're currently vegetarian. You tried pescatarian for a while but returned to vegetarian.''}
\end{casebox}

\begin{casebox}[Case 3: Identity Erosion]
{\small
\textbf{Scenario:} User shares emotionally significant academic milestone.
\textit{Session 12:} ``I finally defended my PhD thesis today! Five years of work on neural memory models.''
\textit{Later sessions:} Casual conversations about weather, movies, daily tasks.
\textit{Query (Session 45):} ``What was my thesis about?''

\textbf{Baseline Failure:} Thesis mention buried under 33 sessions of casual content. Response: ``I don't have information about your thesis.''

\textbf{BMAM Solution:} Amygdala tags the defense announcement as high-salience (emotional significance + milestone event). Protected from forgetting despite low access frequency.

\textbf{Response:} ``Your PhD thesis was on neural memory models. You defended it successfully after five years of research.''}
\end{casebox}

\begin{casebox}[Case 4: Procedural Erosion]
{\small
\textbf{Scenario:} User establishes coding workflow preferences.
\textit{Session 3:} ``Always use TypeScript, never plain JavaScript.''
\textit{Session 7:} ``Format code with Prettier, 2-space indentation.''
\textit{Session 20:} ``I prefer functional components over class components in React.''
\textit{Query:} ``Write me a React component for a login form.''

\textbf{Baseline Failure:} Generates class component with 4-space indentation in JavaScript.

\textbf{BMAM Solution:} Basal ganglia stores procedural preferences as behavioral patterns. Fixed-point detection recognizes consistent preferences across sessions.

\textbf{Response:} Generates functional TypeScript component with Prettier formatting and 2-space indentation.}
\end{casebox}

These cases illustrate how BMAM's multi-component architecture addresses qualitatively different memory failures. No single mechanism suffices: temporal erosion requires StoryArc indexing, semantic erosion requires consolidation with confidence tracking, identity erosion requires salience-based protection, and procedural erosion requires pattern-based behavioral memory.

\section{Prompt Templates}
\label{sec:prompts}

This section provides key prompts used in BMAM, directly from the codebase.

\begin{promptbox}[Query Classification (adaptive\_config.py)]
{\small\ttfamily
Analyze this query and rate each dimension from 0.0 to 1.0:\\[0.3em]
Query: "\{query\}"\\[0.3em]
Dimensions:\\
- temporal: Time/sequence reasoning (when, before, after, order of events)\\
- identity: Personal info recall (my name, my preferences, what I told you)\\
- preference: Choice/comparison (prefer, favorite, which do I like better)\\
- factual: General fact lookup (what is X, define, explain)\\[0.3em]
Return ONLY valid JSON: \{"temporal": 0.X, "identity": 0.X, "preference": 0.X, "factual": 0.X\}
}
\end{promptbox}

\begin{promptbox}[Memory Compression (clean\_agent\_system.py)]
{\small\ttfamily
For the user query "\{query\}", compress the following memories into key facts (3-5 points):\\
1. [memory 1]\\
2. [memory 2]\\
...\\[0.3em]
Keep only core information directly relevant to the query. Be concise.}
\end{promptbox}

\begin{promptbox}[Semantic Consolidation (memory\_coordinator.py)]
{\small\ttfamily
Extract core semantic knowledge from the following \{N\} episodic memories:\\[0.3em]
Date: \{date\}\\[0.3em]
Episodic memories:\\
\{combined\_content\}\\[0.3em]
Please extract:\\
1. Core facts and knowledge points\\
2. Common themes or patterns\\
3. Important entity relationships\\[0.3em]
Output as concise semantic knowledge (2-3 sentences).}
\end{promptbox}

\begin{promptbox}[Context Compaction (context\_compaction.py)]
{\small\ttfamily
Analyze the following conversation history and extract structured notes:\\[0.3em]
\{history\_text\}\\[0.3em]
Output format:\\
1. Core facts: [List 3-5 key pieces of information]\\
2. User preferences: [If any]\\
3. Pending tasks: [Incomplete tasks]\\
4. Key decisions: [Important decisions made]\\
5. Open questions: [Unresolved questions]\\[0.3em]
Requirement: Be extremely concise, keep only the most important information.}
\end{promptbox}

\section{Hyperparameters}
\label{sec:hyperparameters}

Table~\ref{tab:hyperparameters} lists the key hyperparameters used in BMAM experiments.

\begin{table}[h]
\centering
\small
\begin{tabular}{p{0.55\columnwidth}c}
\toprule
Parameter & Value \\
\midrule
\multicolumn{2}{l}{\textit{Brain Region Capacities}} \\
Hippocampus episodic store & 20,000 \\
Temporal lobe semantic store & 70,000 \\
Amygdala salience buffer & 1,000 \\
Prefrontal working memory & 10 \\
Basal ganglia procedural store & 500 \\
\midrule
\multicolumn{2}{l}{\textit{Embedding \& LLM}} \\
Embedding model & text-embed-3-small \\
Embedding dimension & 1536 \\
Response LLM & gpt-4o-mini \\
Judge LLM & gpt-4o-mini \\
Temperature (generation) & 0.7 \\
Temperature (judge) & 0.0 \\
\midrule
\multicolumn{2}{l}{\textit{Pipeline (Algorithm~\ref{alg:bmam-pipeline})}} \\
Max collaborative-loop iter.\ $I_{\max}$ & 3 \\
Temporal confidence threshold $\tau_{\text{temp}}$ & 0.55 \\
Reasoning chain conf.\ threshold & 0.40 \\
LearnableRouter learning rate & 0.05 \\
RRF smoothing constant $k$ & 60 \\
Consolidation importance threshold & 0.60 \\
Consolidation access-count threshold & 2 \\
\bottomrule
\end{tabular}
\caption{BMAM hyperparameters.}
\label{tab:hyperparameters}
\end{table}

\section{Reproducibility Checklist}
\label{sec:reproducibility}

We provide the following information for reproducibility:

\paragraph{Code and Data}
\begin{itemize}
\item The complete BMAM implementation ($\sim$307 Python files) is publicly released at \url{https://github.com/innovation64/BMAM}.
\item Benchmark datasets are publicly available: LoCoMo, LongMemEval, PersonaMem, PrefEval.
\item Evaluation scripts follow the MemOS protocol (repository and version cited in the Evaluation Protocol footnote).
\item No GPU required; reproduction needs only Python\,3.10$+$ and an OpenAI-compatible API key.
\end{itemize}

\paragraph{Random Seeds}
\begin{itemize}
\item Embedding and retrieval are deterministic
\item LLM responses use temperature 0.0 for judge, 0.7 for generation
\item Results may vary slightly across runs due to LLM non-determinism
\end{itemize}

\paragraph{Limitations of Reproducibility}
\begin{itemize}
\item API model versions may change over time
\item Some baseline numbers from MemOS paper; select baselines re-run with official scripts
\item Minor hyperparameter sensitivity not fully characterized
\end{itemize}

\end{document}